\documentclass{article}

\usepackage{microtype}
\usepackage{graphicx}
\usepackage{subcaption}
\usepackage{booktabs} 

\usepackage{hyperref}

\usepackage[preprint]{icml2026}

\usepackage{amsmath}
\usepackage{amssymb}
\usepackage{mathtools}
\usepackage{amsthm}
\usepackage{bbm}

\usepackage[capitalize,noabbrev]{cleveref}

\theoremstyle{plain}

\theoremstyle{definition}

\theoremstyle{remark}

\usepackage[textsize=tiny]{todonotes}

\icmltitlerunning{Training-Free Image Editing with Visual Context Integration and Concept Alignment}

\begin{document}
\twocolumn[
  \icmltitle{Training-Free Image Editing with \\ Visual Context Integration and Concept Alignment}



  \icmlsetsymbol{intern}{$^\dagger$}

  \begin{icmlauthorlist}
    \icmlauthor{Rui Song}{hkust,ali,intern}
    \icmlauthor{Guo-Hua Wang}{ali}
    \icmlauthor{Qing-Guo Chen}{ali}
    \icmlauthor{Weihua Luo}{ali}
    \icmlauthor{Tongda Xu}{tsinghua}
    \icmlauthor{Zhening Liu}{hkust}
    \icmlauthor{Yan Wang}{tsinghua}
    \icmlauthor{Zehong Lin}{lingnan}
    \icmlauthor{Jun Zhang}{hkust}

  \end{icmlauthorlist}


  \icmlaffiliation{hkust}{HKUST}
  \icmlaffiliation{ali}{Alibaba International Digital Commerce Group}
  \icmlaffiliation{tsinghua}{Institute for AI Industry Research (AIR), Tsinghua University}
  \icmlaffiliation{lingnan}{School of Data Science, Lingnan University}

  \icmlcorrespondingauthor{Jun Zhang}{eejzhang@ust.hk}

  \icmlkeywords{Machine Learning, ICML}

  {
    \vspace{0.1em}
    \centering
    \includegraphics[width=0.99\textwidth]{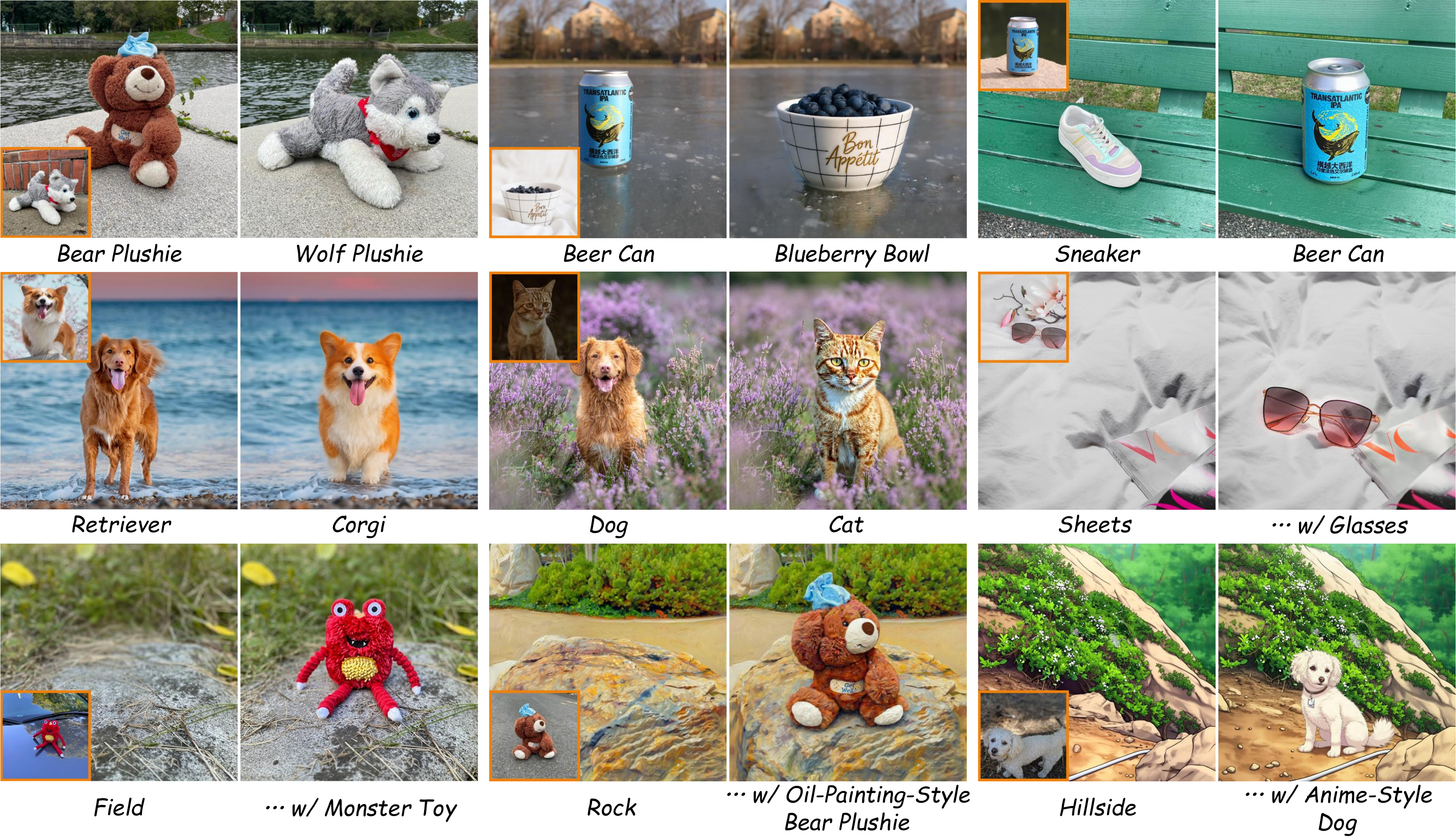}%
    \captionsetup[figure]{hypcap=false}
    \captionof{figure}{Results of the proposed VicoEdit. The left column of each image pair shows the source and context images, while the right column presents the editing result.}
    \label{fig:teaser}
    }

  \vskip 0.1in
]

\printAffiliationsAndNotice{$^\dagger$ Work done during the internship at Alibaba International Digital Commerce Group. Email: $<$rrui.song@connect.ust.hk$>$.}



\begin{abstract}

In image editing, it is essential to incorporate a context image to convey the user's precise requirements, such as subject appearance or image style. Existing training-based visual context-aware editing methods incur data collection effort and training cost. On the other hand, the training-free alternatives are typically established on diffusion inversion, which struggles with consistency and flexibility. In this work, we propose \emph{VicoEdit}, a training-free and inversion-free method to inject the visual context into the pretrained text-prompted editing model. More specifically, VicoEdit directly transforms the source image into the target one based on the visual context, thereby eliminating the need for inversion that can lead to deviated trajectories. Moreover, we design a posterior sampling approach guided by concept alignment to enhance the editing consistency. Empirical results demonstrate that our training-free method achieves even better editing performance than the state-of-the-art training-based models.
\end{abstract}

\section{Introduction}
Image editing aims to modify a source image in accordance with user instructions while preserving the non-target regions. This technology is instrumental to a wide range of creative applications, such as advertising, product design, and personalized content generation. Recent years have witnessed remarkable advancements in image editing techniques \cite{huang2025diffusion}. The majority of editing methods use a text prompt to control editing \cite{mokady2023null, brooks2023instructpix2pix}. However, textual instruction often cannot achieve fine-grained manipulation (e.g., detailed subject appearance or image style change) due to the ambiguity of the language. Thus, it is important to introduce the visual context for fine-grained control \cite{lidreamedit, lu2023tf, shin2025large}.

State-of-the-art visual context-aware editing models \cite{flux-2-2025, wu2025qwen} perform large-scale pretraining using millions of multi-subject image pairs. Each pair consists of a source image, a context image, a textual prompt, and a target image. However, the data curation pipeline is complicated and expensive, typically requiring multiple runs of the image generation model and the vision-language model (VLM) \cite{chen2025xverse, she2025mosaic}. Besides, training a large model is also computation-intensive, which further increases the cost of training-based approaches. 

To avoid expensive data collection and large-scale pretraining, training-free visual context integration methods have been developed \cite{lu2023tf, pham2024tale, li2024tuning}. These methods directly perform context-aware editing based on an image generation model, and they typically use diffusion inversion \cite{songdenoising, mengsdedit} to obtain the noise vectors corresponding to the source and contextual images. These two noise vectors are then merged based on a user-provided mask that indicates the edited regions. The pretrained image generation model is employed to modify the image based on a different caption, starting from the combined noise vector. Nevertheless, the trajectory estimated by diffusion inversion is inaccurate \cite{mokady2023null, ju2024pnp}. Therefore, these methods may fail to preserve details in the source and context images. Furthermore, their reliance on \emph{the user-provided mask} complicates the editing pipeline and weakens the flexibility \cite{hertz2022prompt}.

\vspace{\fill}

To address these challenges, we propose \emph{VicoEdit}, which introduces the \textbf{Vi}sual \textbf{co}ntext to a text-prompted image \textbf{Edit}ing model without training or inversion. VicoEdit combines the velocity fields of the inversion and sampling processes, so that the source image can be directly translated to the target image \cite{kulikov2025flowedit}. Specifically, features from the source image are preserved in the latent embedding, while the visual context is introduced through the attention blocks of the text-prompted editing model. Moreover, we formulate a concept alignment guidance to align the unmodified visual concepts in the source and edited images. These concepts are identified according to the distance between the image and textual concept embeddings, and they will guide the editing through the diffusion posterior sampling \cite{chung2023diffusion}. Besides, VicoEdit is agnostic to the model architecture and does not require user-provided editing masks, significantly enhancing its flexibility. 

Empirically, VicoEdit is implemented on several popular text-driven image editing models \cite{batifol2025flux, wu2025qwen, wang2025ovisu1technicalreport}, and it demonstrates strong performance regarding instruction following and editing faithfulness. Experimental results show that VicoEdit outperforms state-of-the-art training-free and training-based multi-reference image editing models, such as FLUX.2-dev \cite{flux-2-2025} and Qwen-Image-Edit-2511 \cite{wu2025qwen}. Moreover, it yields comparable performance to the latest closed-source commercial image editing models, such as Seedream 5.0 Lite \cite{seedream-5-0-2026} and Nano Banana 2 \cite{nano-banana-2-2026}.

The contributions of this paper are summarized as follows:
\begin{itemize}
    \item We propose VicoEdit, a training-free context-aware image editing method. It achieves superior performance compared to training-based approaches without expensive data collection and pretraining process.
    \item We develop an inversion-free editing pipeline, which improves the editing fidelity and flexibility compared to existing training-free methods.
    \item We devise a concept-alignment method by posterior sampling, to enhance the faithfulness and consistency to the source image.
\end{itemize}

\section{Related Work}
\subsection{Text-Prompted Image Editing}
Existing image editing techniques can be classified into training-based and training-free categories. Training-based methods perform editing based on a pretrained conditional generation model \cite{brooks2023instructpix2pix, huang2024smartedit, deng2025emerging, wu2025omnigen2, wang2025seededit}. These models first encode the textual instruction and the source image into the latent space using the tokenizer \cite{raffel2020exploring, radford2021learning, rombach2022high} or VLM \cite{liu2023visual, bai2025qwen2}. Then, textual and visual embeddings are integrated as the condition for generation \cite{li2023blip, esser2024scaling}. On the other hand, training-free image editing methods generalize the pretrained text-to-image model to image editing, where the most widely used techniques are diffusion inversion \cite{mengsdedit,kim2022diffusionclip,wang2025taming,rout2024beyond,rout2025semantic} and attention manipulation \cite{hertz2022prompt, cao2023masactrl}. In contrast to these text-prompted editing methods, this paper explores editing the image conditioned on both the text instruction and another visual context image to apply more fine-grained control.

\subsection{Multi-Reference Image Editing}
Multi-reference image editing aims to generate images conditioned on a text prompt and multiple context images, and state-of-the-art models requires training on large-scale datasets \cite{wang2025ms, mou2025dreamo, wu2025qwen, flux-2-2025}. Although it provides an effective solution for visual context integration, the training process requires heavy computations since it has to process features of multiple condition images. Furthermore, data curation is also computationally expensive, due to the involvement of numerous large models. For example, a commonly used pipeline \cite{chen2025xverse,she2025mosaic,wu2025less} starts with images that include multiple subjects. These images are used as the target for generation, and a VLM is employed to write the corresponding captions. Then, subjects in the target image are grounded \cite{liu2024grounding}, segmented \cite{ravi2025sam}, and refined \cite{flux2024} to generate the context images. Finally, a VLM or vision foundation model \cite{oquab2024dinov2} is employed for data filtering. The high cost of data collection and training motivates us to develop a training-free visual context integration method for editing tasks.

Context-driven editing can also be achieved using the training-free diffusion inversion methods \cite{lu2023tf, pham2024tale}. Given a source image $\boldsymbol{x}^{src}_{0}$ and a context image $\boldsymbol{x}^{ctx}_{0}$, these methods first derive their corresponding noise vectors $\boldsymbol{x}^{src,*}_{T}$ and $\boldsymbol{x}^{ctx,*}_{T}$. Subsequently, these two noise vectors are merged as $\boldsymbol{x}^{*}_{T} = (\boldsymbol{1}-\boldsymbol{m}) \boldsymbol{x}^{src,*}_{T} + \boldsymbol{m} \boldsymbol{x}^{ctx,*}_{T}$, where $\boldsymbol{m}$ is a user-given mask that specifies the regions to be edited. Finally, the edited image is generated from $\boldsymbol{x}^{*}_{T}$, based on the caption of the desired image. However, a minor error is introduced in each inversion step, and the classifier-free guidance \cite{ho2022classifier} further amplifies this error. As a result, these methods struggle to preserve the details of $\boldsymbol{x}^{src}_{0}$ and $\boldsymbol{x}^{ctx}_{0}$. Moreover, these methods must receive the mask $\boldsymbol{m}$ as an additional input, which complicates the workflow and hinders the usability \cite{hertz2022prompt}. To address these issues, we develop an inversion-free and mask-free method to integrate the visual context features.

\section{Preliminaries}
\subsection{Rectified Flow}
Flow matching \cite{lipman2023flow} transfers a sample from the source distribution $\boldsymbol{z}_1 \sim \pi_1$ (e.g., standard Gaussian) to the target distribution $\boldsymbol{z}_0 \sim \pi_0$ by learning a flow $\boldsymbol{f}$ to predict the velocity field. 
This velocity field serves as the solution of an ordinary differential equation (ODE) $d \boldsymbol{z}_t = \boldsymbol{f}(\boldsymbol{z}_t,t) dt$,
from which one can finally obtain samples from the desired distribution $\pi_0$. In particular, rectified flow \cite{liu2023flow} defines the transformation following the straight path between $\boldsymbol{z}_1$ and $\boldsymbol{z}_0$, i.e., $\boldsymbol{z}_t = (1-t) \boldsymbol{z}_0 + t \boldsymbol{z}_1$. This formulation reduces the transportation cost and sampling steps, making the rectified flow a prominent technique for image generation \cite{esser2024scaling, flux2024}.

\subsection{FlowEdit}
\begin{figure*}[ht!]
\centering
\includegraphics[width=0.99\textwidth]{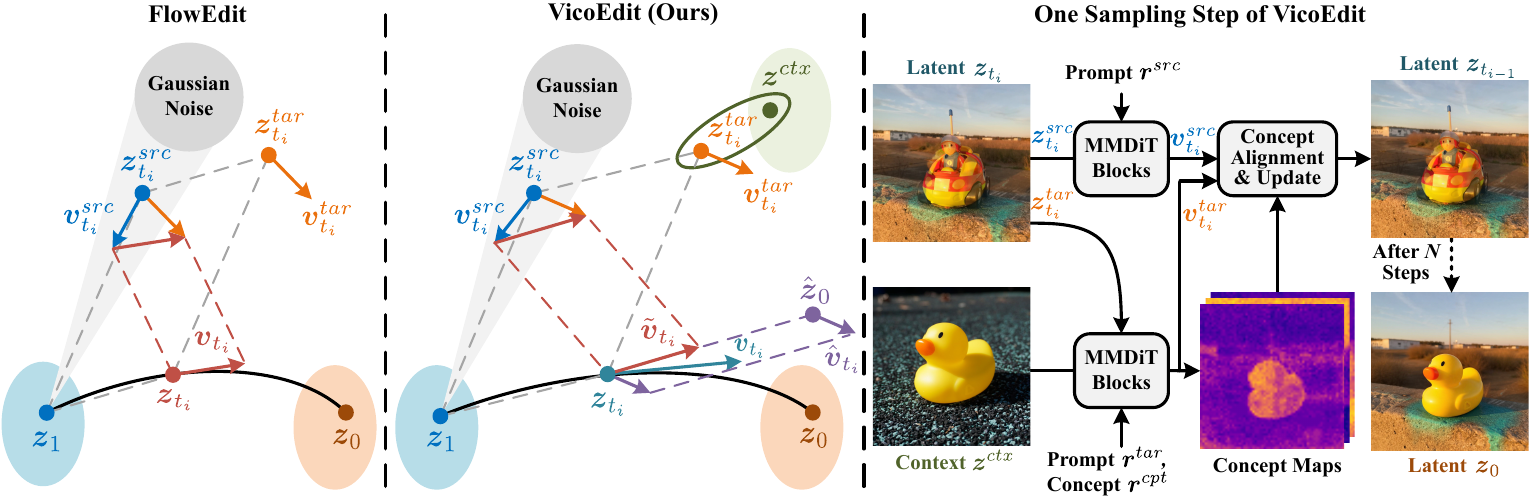}
\caption{The left figure shows the pipeline of FlowEdit. The middle figure illustrates the latent vectors, velocity fields, and sampling trajectory of VicoEdit. The right figure shows the procedure of each sampling step of VicoEdit.}
\label{fig:pipeline}
\end{figure*}

FlowEdit \cite{kulikov2025flowedit} introduces an inversion-free pipeline for text-driven image editing, which delivers better structure preservation capability. It turns out that the inverse and sampling processes can be reinterpreted as a single trajectory by combining their velocity fields. Specifically, it initializes the ODE as the latent of the source image, i.e., $\boldsymbol{z}_1 = \boldsymbol{z}^{src}$. At each timestep, the velocity field of the inverse process is computed as follows:
\begin{align}
    \boldsymbol{v}^{src}_{t_i} = \boldsymbol{f}(\boldsymbol{z}^{src}_{t_i}, \boldsymbol{r}^{src}, {t_i}), \text{ where } \boldsymbol{z}^{src}_{t_i} = (1-{t_i}) \boldsymbol{z}_{1} + {t_i} \boldsymbol{\epsilon}.
    \label{eq:flowedit-src}
\end{align}
In Eq. \ref{eq:flowedit-src}, $\boldsymbol{\epsilon}$ is sampled from the standard Gaussian distribution and $\boldsymbol{r}^{src}$ is the caption of the source image. Then, the velocity field for the sampling process is formulated as:
\begin{align}
    \boldsymbol{v}^{tar}_{t_i} = \boldsymbol{f}(\boldsymbol{z}^{tar}_{t_i}, \boldsymbol{r}^{tar}, {t_i}), \text{ where } \boldsymbol{z}^{tar}_{t_i} = \boldsymbol{z}_{t_i} + \boldsymbol{z}^{src}_{t_i} - \boldsymbol{z}_1.
    \label{eq:flowedit-tar}
\end{align}
Here, $\boldsymbol{r}^{tar}$ indicates the caption of the desired image. $\boldsymbol{z}^{tar}_{t_i}$ approximates the latent in the sampling trajectory at timestep $t_i$, which is obtained by moving the latent $\boldsymbol{z}_{t_i}$ along the direction of $\boldsymbol{z}^{src}_{t_i} - \boldsymbol{z}_1$, as illustrated in Fig. \ref{fig:pipeline} (left).
Finally, $\boldsymbol{z}_{t_i}$ is updated by integrating $\boldsymbol{v}^{src}_{t_i}$ and $\boldsymbol{v}^{tar}_{t_i}$:
\begin{align}
\boldsymbol{z}_{t_{i-1}} &= \boldsymbol{z}_{t_i} + (t_{i-1} - t_i) \boldsymbol{v}_{t_i},  \notag\\
\text{with } \boldsymbol{v}_{t_i} &= \mathbb{E}_{\boldsymbol{\epsilon}}[(\boldsymbol{v}^{tar}_{t_i} - \boldsymbol{v}^{src}_{t_i})|\boldsymbol{z}_0],
\end{align}
where the expectation can be estimated by Monte Carlo sampling. In this manner, $\boldsymbol{z}_{t_i}$ traverses a direct path from the source to the target distribution, therefore reducing the transportation cost and improving the editing faithfulness.

\section{VicoEdit: Training-Free Context-Aware Image Editing}
VicoEdit edits the source image $\boldsymbol{x}^{src}$ conditioned on the textual prompt $\boldsymbol{r}$ and another context image $\boldsymbol{x}^{ctx}$. Its pipeline is illustrated in Fig. \ref{fig:pipeline}, where $\boldsymbol{z}$ denotes the latent representation of $\boldsymbol{x}$. Overall, VicoEdit maintains the features of $\boldsymbol{x}^{src}$ within the latent $\boldsymbol{z}_t$, and contextual features $\boldsymbol{z}^{ctx}$ are injected to $\boldsymbol{z}_t$ through the pretrained multi-modal diffusion transformer (MMDiT) blocks \cite{esser2024scaling}. At the $i$-th timestep $t_i$, it couples the inverse and the sampling velocity fields into a unified velocity field $\tilde{\boldsymbol{v}}_{t_i}$. This allows for a direct trajectory from the initial latent $\boldsymbol{z}_1 =\boldsymbol{z}^{src}$ to the target-domain latent $\boldsymbol{z}_0$. In addition, concept alignment generates a guidance term $\hat{\boldsymbol{v}}_{t_i}$ to ensure the consistency between the source and edited images.

\subsection{Visual Context Integration}
\label{sec:visual-context-integration}
Inspired by FlowEdit, we propose a training-free method to integrate the visual context into the text-prompted editing model. In each step, the proposed method first computes the intermediate latents $\boldsymbol{z}^{src}_{t_i}$ and $\boldsymbol{z}^{tar}_{t_i}$ for the inverse and sampling processes using Eq. \ref{eq:flowedit-src} and Eq. \ref{eq:flowedit-tar}. Then, it predicts the corresponding velocity fields. The inverse velocity field $\boldsymbol{v}^{src}_{t_i}$ corresponds to the flow that generates the source image $\boldsymbol{x}^{src}$. Since the generation of $\boldsymbol{x}^{src}$ depends only on the textual prompt $\boldsymbol{r}^{src}$, the inverse velocity field is estimated by $\boldsymbol{v}^{src}_{t_i} = \boldsymbol{f}(\boldsymbol{z}^{src}_{t_i}, \boldsymbol{r}^{src}, t_i)$, without introducing the visual context features $\boldsymbol{z}^{ctx}$.
On the other hand, we aim to generate $\boldsymbol{z}_0$ conditioned on both the textual prompt $\boldsymbol{r}^{tar}$ and the visual context $\boldsymbol{z}^{ctx}$. To this end, $\boldsymbol{z}^{tar}_{t_i}$ is aggregated with $\boldsymbol{z}^{ctx}$ when predicting the sampling velocity field:
\begin{align}
 \boldsymbol{v}^{tar}_{t_i} = \boldsymbol{f}(\boldsymbol{z}^{tar}_{t_i}, \boldsymbol{r}^{tar}, \boldsymbol{z}^{ctx}, t_i),
\end{align}
where $\boldsymbol{z}^{tar}_{t_i}$, $\boldsymbol{r}^{tar}$, and $\boldsymbol{z}^{ctx}$ serve as the noise, textual condition, and visual condition tokens in the MMDiT blocks, respectively. Within the pretrained editing model, the attention layers in the MMDiT integrate features of the generated image (i.e., noise tokens) and the visual context. As $\boldsymbol{z}^{tar}_{t_i}$ approximates the sample in the reverse path at timestep $t_i$, it makes sense to aggregate features of $\boldsymbol{z}_t$ and $\boldsymbol{z}^{ctx}$ by regarding $\boldsymbol{z}^{tar}_{t_i}$ as noise tokens. Finally, the combined velocity field is given by $\tilde{\boldsymbol{v}}_{t_i} = \mathbb{E}[\boldsymbol{v}^{tar}_{t_i} - \boldsymbol{v}^{src}_{t_i}|\boldsymbol{z}_0]$. Crucially, our method directly extends the pretrained text-prompted editing model to context-aware editing, without requiring any additional training or fine-tuning.

\begin{figure}[t]
\centering
\includegraphics[width=0.48\textwidth]{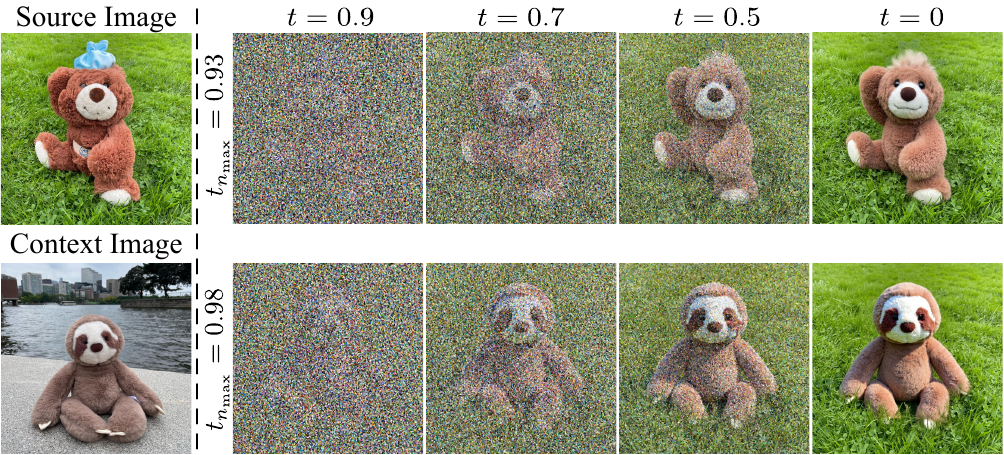}
\caption{Visualization of $\boldsymbol{z}^{tar}_t$ at different timesteps. We visualize the latents from two different trajectories, where the timesteps for starting sampling (i.e., $t_{n_\text{max}}$) are $0.93$ and $0.98$, respectively. The model is instructed to replace the bear with the sloth. The visualization verifies that global features are generated at early steps, and skipping the early stage fails to alter the subject appearance.}
\label{fig:n_max}
\end{figure}

Furthermore, we reformulate the sampling strategy to strengthen the influence of the visual context $\boldsymbol{z}^{ctx}$. Text-prompted editing methods often skip a few initial sampling steps to ensure the faithfulness \cite{mengsdedit, kulikov2025flowedit}. However, existing works \cite{hoogeboom2023simple, chen2024training} and our empirical studies show that the image outline and global features of subjects are commonly generated at early steps. As shown in Fig. \ref{fig:n_max}, when these steps are omitted, the model may fail to make significant changes to the image according to $\boldsymbol{z}^{ctx}$. To address this issue, we start sampling at the timestep $t_{n_\text{max}}$ that is close to $1$. Although this strategy is effective, it is still inadequate to yield high-quality results. An underlying reason is that the prediction of $\tilde{\boldsymbol{v}}_{t_i}$ tends to be unstable at high noise levels (i.e., when $t$ approaches $1$). To overcome this challenge, we collect $K$ samples of $\boldsymbol{z}^{src}_{t_i}$ and $\boldsymbol{z}^{tar}_{t_i}$:
\begin{align}
    \boldsymbol{z}^{src}_{t_i, k} &= (1-{t_i}) \boldsymbol{z}_{1} + {t_i} \boldsymbol{\epsilon}_k, \\
    \boldsymbol{z}^{tar}_{t_i,k} &= \boldsymbol{z}_{t_i} + \boldsymbol{z}^{src}_{t_i,k} - \boldsymbol{z}_1.
\end{align}
Then, we compute and average the corresponding velocity fields as $\tilde{\boldsymbol{v}}_{t_i} = \frac{1}{K} \sum_{k=1}^{K}(\boldsymbol{v}^{tar}_{t_i,k} - \boldsymbol{v}^{src}_{t_i,k})$ to stabilize the estimation of the expectation.

\begin{figure}[t]
\centering
\includegraphics[width=0.48\textwidth]{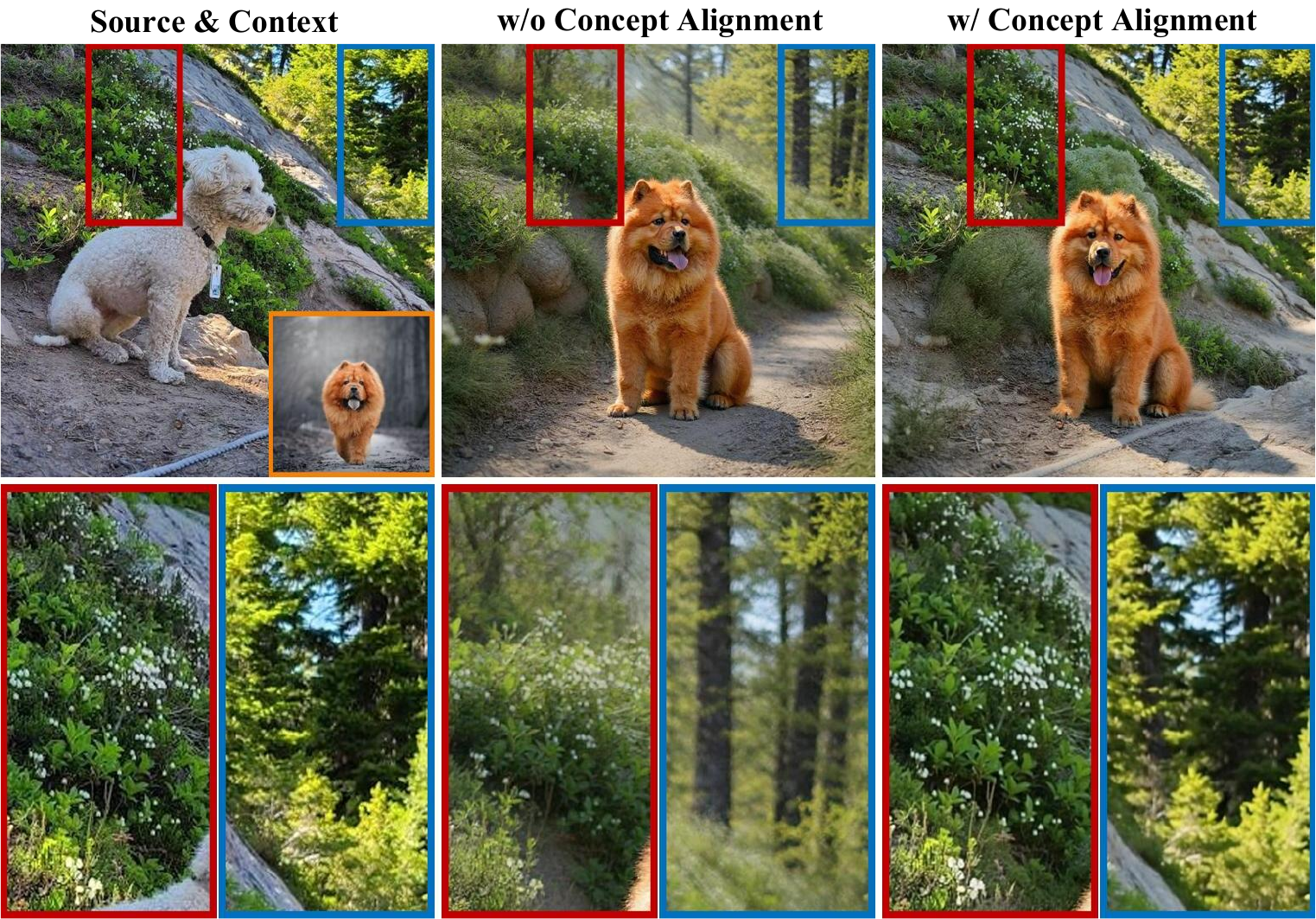}
\caption{Editing results with or without concept alignment. Concept alignment preserves details in the source image.}
\label{fig:dps_visualization}
\end{figure}

\subsection{Concept Alignment}
Although the inversion-free visual context integration improves the editing consistency, it may still fail to restore the detailed patterns accurately, as shown in Fig. \ref{fig:dps_visualization}. Besides, we set a relatively large $t_{n_{\text{max}}}$ to emphasize the impact of $\boldsymbol{z}^{ctx}$, which somewhat compromises the faithfulness. Regarding these issues, we develop a concept alignment approach to improve the fidelity to the source image.

\subsubsection{Concept Classification}
Concept alignment matches the unmodified regions in the source and edited images, without affecting the edited regions. Existing works have shown that the attention score between text and image tokens reflects the correspondence between semantic concepts and pixels \cite{hertz2022prompt, chefer2023attend}. This motivates us to recognize the unmodified regions using the text-to-image attention score. Specifically, the concept alignment module receives several concept words $\boldsymbol{c} = [\boldsymbol{c}_{pos}, \boldsymbol{c}_{neg}]$, where $\boldsymbol{c}_{pos}$ and $\boldsymbol{c}_{neg}$ indicate the concepts to be preserved and modified, respectively. Then, $\boldsymbol{c}$ is encoded into the embedding $\boldsymbol{r}^{cpt}$ and propagated together with other tokens using the ConceptAttention algorithm \cite{helbling2025conceptattention}. In particular, for the $m$-th MMDiT block at timestep $t$, the concept tokens are attended to image tokens as:
\begin{align}
    \boldsymbol{r}^{cpt}_{t, m+1} = \text{Attn}(\boldsymbol{q}^{cpt}_{t,m}, [\boldsymbol{k}^{cpt}_{t,m}, \boldsymbol{k}^{img}_{t,m}], [\boldsymbol{v}^{cpt}_{t,m}, \boldsymbol{v}^{img}_{t,m}]).
\end{align}
Here, $\boldsymbol{q}^{cpt}_{t,m}$, $\boldsymbol{k}^{cpt}_{t,m}$, and $\boldsymbol{v}^{cpt}_{t,m}$ are generated from $\boldsymbol{r}^{cpt}_{t,m}$, whereas $\boldsymbol{k}^{img}_{t,m}$ and $\boldsymbol{v}^{img}_{t,m}$ are extracted from $\boldsymbol{z}^{tar}_{t,m}$. The categorical distribution $\boldsymbol{d}$, which assigns image tokens to specific concept classes, is calculated based on the inner product between the image and concept embeddings:
\begin{align}
    \boldsymbol{d}_{t,m} = \text{softmax}(\boldsymbol{z}^{tar}_{t,m+1} \cdot(\boldsymbol{r}^{cpt}_{t, m+1})^T ),
\end{align}
where the softmax operator is applied on the category dimension. Afterwards, $\boldsymbol{d}_{t,m}$ from all DiT blocks and $K$ samples of $\boldsymbol{z}^{tar}_t$ are aggregated by averaging, which results in $\boldsymbol{d}_{t}$. Finally, we interpolate $\boldsymbol{d}_t$ to the resolution of $\boldsymbol{x}^{src}$ and calculate a binary mask $\boldsymbol{m}_t$ to indicate whether a pixel belongs to the preserved concepts as $\boldsymbol{m}_t = 1\Big(\sum_{c \in \boldsymbol{c}^{pos}} \boldsymbol{d}_{t,c} \geq \tau \Big)$. Here, $1(\cdot)$ is the indicator function, $\boldsymbol{d}_{t,c}$ denotes the probability of being the concept $c$, and $\tau$ is the threshold. 

\begin{algorithm}[t]
  \caption{VicoEdit}
  \label{algorithm}
  \begin{algorithmic}
    \STATE {\bfseries Input:} $\boldsymbol{z}^{src}$, $\boldsymbol{z}^{ctx}$, $\boldsymbol{r}^{src}$, $\boldsymbol{r}^{tar}$, $\boldsymbol{r}^{cpt}$, $\left \{t_i \right \}_{i=1}^N$, \\ $\quad \quad \quad n_{\text{max}}$, $K$, $\tau$, $\left\{\alpha_{t_i} \right \}_{i=1}^{N}$, $\sigma$
    \STATE {\bfseries Initialize:} $\boldsymbol{z}_1 = \boldsymbol{z}^{src}$.
    \FOR{$i=n_{\text{max}}$ {\bfseries to} $1$}
    \FOR{$k=1$ {\bfseries to} $K$}
    \STATE $\boldsymbol{\epsilon}_k \sim \mathcal{N}(\boldsymbol{0}, \boldsymbol{I})$
    \STATE $\boldsymbol{z}^{src}_{t_i,k} \gets (1-t_i) \boldsymbol{z}_1 + t_i \boldsymbol{\epsilon}_k$
    \STATE $\boldsymbol{z}^{tar}_{t_i,k} \gets \boldsymbol{z}_{t_i} + \boldsymbol{z}^{src}_{t_i,k} - \boldsymbol{z}_1$
    \STATE $\boldsymbol{v}^{src}_{t_i,k} \gets \boldsymbol{f}(\boldsymbol{z}^{src}_{t_i,k}, \boldsymbol{r}^{src}, t_i)$
    \STATE $\boldsymbol{v}^{tar}_{t_i,k}, \boldsymbol{d}_{t_i, k} \gets \boldsymbol{f}(\boldsymbol{z}^{tar}_{t_i,k}, \boldsymbol{z}^{ctx}, \boldsymbol{r}^{tar}, \boldsymbol{r}^{cpt}, t_i)$
    \ENDFOR
    \STATE $\tilde{\boldsymbol{v}}_{t_i} \gets \frac{1}{K} \sum_k (\boldsymbol{v}^{tar}_{t_i,k} - \boldsymbol{v}^{src}_{t_i,k})$
    \STATE $\boldsymbol{d}_{t_i} \gets \frac{1}{K} \sum_k \boldsymbol{d}_{t_i,k}$
    \STATE $\boldsymbol{m}_{t_i} \gets \mathbbm{1}[\sum_{c \in \boldsymbol{c}^{pos}} \boldsymbol{d}_{t_i,c} \geq \tau]$
    \STATE $\boldsymbol{s}, \tilde{\boldsymbol{s}} \sim \mathcal{N}(\boldsymbol{0}, \sigma^2 \boldsymbol{I})$
    \STATE $\boldsymbol{y} \gets \boldsymbol{m}_{t_i} \boldsymbol{x}_0 + \boldsymbol{s}$
    \STATE $\hat{\boldsymbol{z}}_{0} \gets \boldsymbol{z}_{t_i} - t_i \tilde{\boldsymbol{v}}_{t_i}$
    \STATE $\hat{\boldsymbol{v}}_{t_i} \gets \alpha_{t_i} \nabla_{\boldsymbol{z}_{t_i}} || \boldsymbol{y} - (\boldsymbol{m}_{t_i} \mathcal{D}(\hat{\boldsymbol{z}}_{0}) + \tilde{\boldsymbol{s}})||^2_2$
    \STATE $\boldsymbol{z}_{t_{i-1}} \gets \boldsymbol{z}_{t_i} + (t_{i-1} - t_i) (\tilde{\boldsymbol{v}}_{t_i} + \hat{\boldsymbol{v}}_{t_i})$
    \ENDFOR
    \STATE {\bfseries Return: $\mathcal{D}(\boldsymbol{z}_0)$}
  \end{algorithmic}
\end{algorithm}

\subsubsection{Concept-Guided Posterior Sampling}
After establishing the correspondence between the pixels and the concepts, we use the diffusion posterior sampling (DPS) \cite{chung2023diffusion} to preserve the structures of $\boldsymbol{x}_0$ in the unmodified regions. Formally, our target is to sample from the posterior distribution $p(\boldsymbol{z}|\boldsymbol{y})$, where $\boldsymbol{y}$ indicates the unchanged regions in $\boldsymbol{x}_0$:
\begin{align}
    \boldsymbol{y} = \boldsymbol{m}_t \boldsymbol{x}_0 + \boldsymbol{s}, \quad \text{where } \boldsymbol{s} \sim \mathcal{N}(\boldsymbol{0}, \sigma^2 \boldsymbol{I}).
\label{eq:dps-measurement}
\end{align}
To achieve this, we introduce an additional guidance:
\begin{align}
\hat{\boldsymbol{v}}_t = \nabla_{\boldsymbol{z}_t} \log p (\boldsymbol{y}|\boldsymbol{z}_t).
\end{align}
By combining the unconditional velocity field and the guidance $\nabla_{\boldsymbol{z}_t} \log p (\boldsymbol{y}|\boldsymbol{z}_t)$, rectified flow can sample from the posterior distribution $p(\boldsymbol{z}|\boldsymbol{y})$. Specifically, the rectified flow defined on a Gaussian path generates samples from $p(\boldsymbol{z}_0|\boldsymbol{y})$ at $t=0$ by solving the ODE using the velocity field 
\begin{align}
u_t(\boldsymbol{z}_t|\boldsymbol{y}) = u_t(\boldsymbol{z}_t) + b_t \nabla_{\boldsymbol{z}_t} \log p (\boldsymbol{y}|\boldsymbol{z}_t),
\label{eq:velocity-decomposition}
\end{align}
where $u_t(\boldsymbol{z}_t)$ is the unconditional velocity field, and $b_t = -\frac{t}{1-t}$ \cite{dao2023flow}. Intuitively, $\hat{\boldsymbol{v}}_t$ plays a similar role as the classifier guidance \cite{dhariwal2021diffusion}.  

Note that Eq. \ref{eq:velocity-decomposition} is strictly valid for Gaussian probability flow, while VicoEdit does not evolve samples along the Gaussian path. However, we find it effective to \emph{heuristically} extend Eq. \ref{eq:velocity-decomposition} to the VicoEdit framework. Suppose $\boldsymbol{v}_t(\boldsymbol{z}_t|\boldsymbol{y})$ generates a conditional probability path that directly connects the source and target domains. We decompose $\boldsymbol{v}_t(\boldsymbol{z}_t|\boldsymbol{y})$ into the unconditional velocity field $\tilde{\boldsymbol{v}}_t$ and the classifier guidance $\nabla_{\boldsymbol{z}_t} \log p (\boldsymbol{y}|\boldsymbol{z}_t)$. As $\tilde{\boldsymbol{v}}_t$ has been modeled in Sec. \ref{sec:visual-context-integration}, our subsequent target is to estimate $\nabla_{\boldsymbol{z}_t} \log p (\boldsymbol{y}|\boldsymbol{z}_t)$ using the DPS algorithm. DPS shows that the likelihood function $p(\boldsymbol{y}|\boldsymbol{x}_t)$ can be approximated as:
\begin{align}
p&(\boldsymbol{y}|\boldsymbol{x}_t) \approx p(\boldsymbol{y}|\hat{\boldsymbol{x}}_{0}), \quad \text{where } \hat{\boldsymbol{x}}_{0} = \mathbb{E}[\boldsymbol{x}_{0}|\boldsymbol{x}_t].
\label{eq:dps}
\end{align}
Here, $\hat{\boldsymbol{x}}_{0}$ indicates the expectation of the clean image $\boldsymbol{x}_0$ given the noisy one $\boldsymbol{x}_t$, which can be obtained by applying Tweedie's formula \cite{efron2011tweedie}. Considering that $p(\boldsymbol{y}|\boldsymbol{x}_0)$ is a Gaussian distribution (as defined in Eq.\ref{eq:dps-measurement}), $\nabla_{\boldsymbol{x}_t} \log p (\boldsymbol{y}|\boldsymbol{x}_t)$ is given by:
\begin{align}
\nabla_{\boldsymbol{x}_t} \log p (\boldsymbol{y}|\boldsymbol{x}_t) \approx -\frac{1}{\sigma^2} \nabla_{\boldsymbol{x}_t} || \boldsymbol{y} - (\boldsymbol{m}_t \hat{\boldsymbol{x}}_{0} + \tilde{\boldsymbol{s}})||^2_2,
\label{eq:dps-gradient}
\end{align}
where $\tilde{\boldsymbol{s}}$ is sampled from $\mathcal{N}(\boldsymbol{0}, \sigma^2 \boldsymbol{I})$. More details about DPS are presented in Appendix \ref{sec:appendix-dps}.

\begin{figure}[t]
\centering
\includegraphics[width=0.47\textwidth]{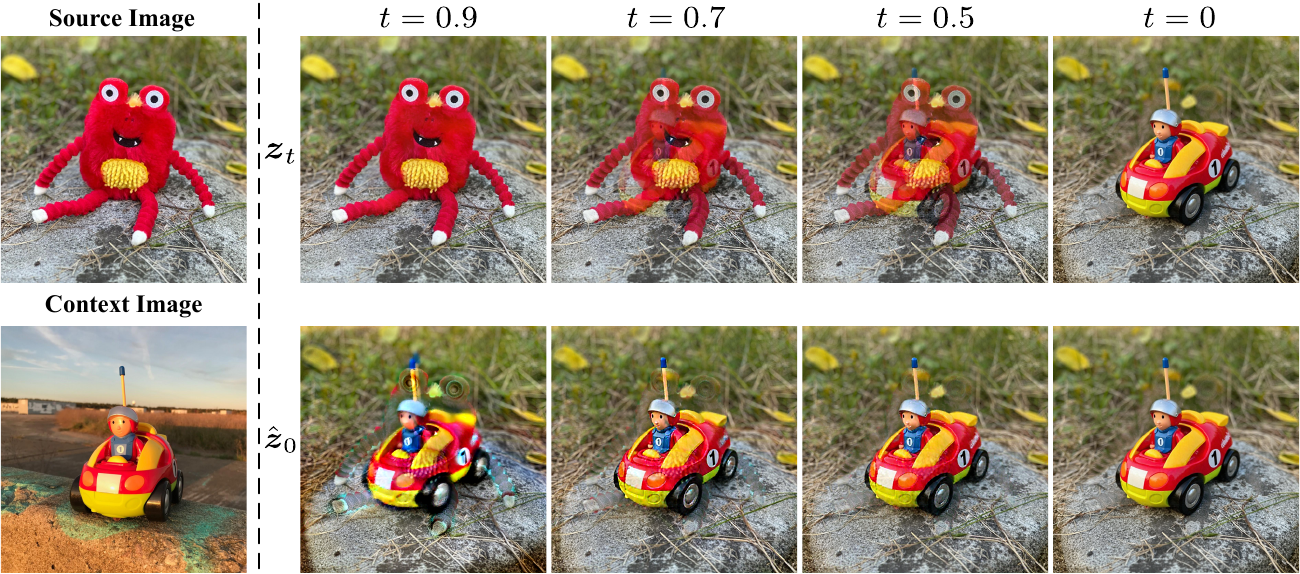}
\caption{Visualization of $\boldsymbol{z}_t$ and $\hat{\boldsymbol{z}}_0$ at different timesteps. Concept alignment guidance accurately predicts $\boldsymbol{z}_0$ even at early timesteps (e.g., when $t=0.9$).}
\label{fig:preview-visualization}
\end{figure}

As shown in \cite{rout2023solving}, DPS can be extended to the latent space as:
\begin{align}
    \nabla_{\boldsymbol{z}_t} \log p(\boldsymbol{y}|\boldsymbol{z}_t) \approx \nabla_{\boldsymbol{z}_t} \log p(\boldsymbol{y}|\hat{\boldsymbol{x}}_0 = \mathcal{D}(\mathbb{E}[\boldsymbol{z}_0|\boldsymbol{z}_t])),
\label{eq:latent-dps}
\end{align}
where $\mathcal{D}$ is the VAE decoder. Besides, this paper \cite{rout2023solving} proposes another latent DPS formulation that provides better theoretical properties. However, we use this vanilla extension form because it performs well in practice while saving computational overhead compared to other formulations. Using Tweedie's formula, we can estimate the posterior expectation of the rectified flow:
\begin{align}
    \hat{\boldsymbol{z}}_0 := \mathbb{E}[\boldsymbol{z}_0|\boldsymbol{z}_t] = \boldsymbol{z}_t -t u_t(\boldsymbol{z}_t).
\label{eq:rf-tweedie}
\end{align}
It should be clarified that the estimation of Eq. \ref{eq:rf-tweedie} necessitates that $p(z_t|z_0)$ follows a Gaussian distribution, which is not guaranteed in the case of VicoEdit. Nevertheless, we show that we can use a similar formulation to estimate $\hat{\boldsymbol{z}}_0$ based on $\boldsymbol{z}_t$. 
In the context of inversion-free editing, $\boldsymbol{z}^{src}_t$ and $\boldsymbol{z}^{tar}_t$ approximate the samples in the forward and reverse processes, respectively. Therefore, it is reasonable to assume that $p(z^{src}_t|z_1)$ and $p(z^{tar}_t|z_0)$ are Gaussian distributions, and hence the posterior expectation is given by:
\begin{align}
    \hat{\boldsymbol{z}}_0 \approx \boldsymbol{z}^{tar}_t -t \boldsymbol{v}^{tar}_t, \quad \boldsymbol{z}_1 \approx z^{src}_t - t\boldsymbol{v}^{src}_t.
\label{eq:tweedies-for-z0-and-z1}
\end{align}
Then, $\hat{\boldsymbol{z}}_0$ can be approximated as:
\begin{align}
    \hat{\boldsymbol{z}}_0 &\approx \boldsymbol{z}^{tar}_t -t \boldsymbol{v}^{tar}_t \notag\\
    &= \boldsymbol{z}_t + \boldsymbol{z}^{src}_t - \boldsymbol{z}_1 - t \boldsymbol{v}^{tar}_t \notag \\
    &\approx \boldsymbol{z}_t + \boldsymbol{z}^{src}_t - (z^{src}_t - t\boldsymbol{v}^{src}_t) - t \boldsymbol{v}^{tar}_t \notag \\
    &=\boldsymbol{z}_t - t (\boldsymbol{v}^{tar}_t - \boldsymbol{v}^{src}_t) \notag \\
    &=\boldsymbol{z}_t - t\tilde{\boldsymbol{v}}_t.
\label{eq:hat-z0}
\end{align}
Here, the first and third step approximates $z_0$ and $z_1$ with Tweedie's estimation (Eq. \ref{eq:tweedies-for-z0-and-z1}). Overall, Eq. \ref{eq:hat-z0} is equivalent to estimate the target-domain latent $\hat{\boldsymbol{z}}_{0}$ using the one-step denoising. Fig. \ref{fig:preview-visualization} visualizes $\hat{\boldsymbol{z}}_0$ at different timesteps by decoding the corresponding latent through the VAE decoder. It is shown that $\hat{\boldsymbol{z}}_0$ provides an accurate prediction of $\boldsymbol{z}_0$ even at early timesteps, which verifies the effectiveness of the approximation in Eq. \ref{eq:hat-z0}. Next, $\hat{\boldsymbol{z}}_{0}$ is projected back to the pixel space through the VAE decoder $\mathcal{D}(\cdot)$, which produces the approximation of $\hat{\boldsymbol{x}}_0$. Finally, plugging Eq. \ref{eq:dps-measurement} into Eq. \ref{eq:latent-dps}, we derive the concept alignment guidance:
\begin{align}
    \nabla_{\boldsymbol{z}_t} \log p(\boldsymbol{y}|\boldsymbol{z}_t) \approx -\frac{1}{\sigma^2} \nabla_{\boldsymbol{z}_t}|| \boldsymbol{y} - (\boldsymbol{m}_t \mathcal{D}(\hat{\boldsymbol{z}}_{0}) + \tilde{\boldsymbol{s}})||^2_2,
    \label{eq:dps-gradient-latent}
\end{align}
where $\hat{\boldsymbol{z}}_0$ is given by Eq. \ref{eq:hat-z0}. An intuitive explanation of Eq. \ref{eq:dps-gradient-latent} is to minimize the difference between the unchanged regions in $\boldsymbol{x}_0$ and $\hat{\boldsymbol{x}}_0$, by updating $\boldsymbol{z}_t$ during sampling.

Combining the inversion-free visual context integration and the concept alignment guidance, the workflow of VicoEdit is presented in Algorithm \ref{algorithm}. In particular, the coefficients $-1/\sigma^2$ in Eq. \ref{eq:dps-gradient-latent} and $b_t$ in Eq. \ref{eq:velocity-decomposition} are absorbed into the hyper-parameter $\alpha_t$, which controls the strength of the concept alignment guidance.

\section{Experiment}
\subsection{Setup}
VicoEdit is implemented upon state-of-the-art text-prompted image editing models, including FLUX.1-Kontext-dev \cite{batifol2025flux}, Qwen-Image-Edit \cite{wu2025qwen}, and Ovis-U1 \cite{wang2025ovisu1technicalreport}. We compare VicoEdit with the training-free approach Diptych Prompting \cite{shin2025large}, training-based multi-reference editing models FLUX.2-dev \cite{flux-2-2025} and Qwen-Image-Edit-2511 (Qwen-2511) \cite{wu2025qwen}, as well as the closed-source commercial editing models Seedream 5.0 Lite \cite{seedream-5-0-2026} and Nano Banana 2 \cite{nano-banana-2-2026}. Notably, VicoEdit is established on the earliest version of Qwen-Image-Edit, which supports only text-prompted editing. More details are specified in the Appendix \ref{sec:appendix-vicoedit-settings} and \ref{sec:appendix-baseline-settings}.

Experiments are conducted on the DreamBooth dataset \cite{ruiz2023dreambooth}. We manually select images from this dataset to form suitable source and context image pairs. These pairs are then processed by FLUX.2 to generate test samples for three different tasks:
\begin{itemize}
\item \textbf{In-domain replacement}: Models are instructed to substitute the subject in the source image with a reference object provided in the context image. 

\item \textbf{In-domain add}: Models are required to insert a certain subject in the context image into the source image, with the latter serving as the background image.

\item \textbf{Cross-domain add}: Models need to first transfer the style of the subject in the context image, and then harmonize the stylized subject with the source image to create a coherent composition.

\end{itemize}
The resultant evaluation dataset consists of over 300 image pairs. Please refer to Appendix \ref{sec:appendix-dataset} for more details about dataset curation.

We use LPIPS \cite{zhang2018unreasonable} to quantify the consistency between the source and edited images. The similarity between the subjects in the context and edited images is measured by DINO similarity \cite{ruiz2023dreambooth, oquab2024dinov2}. Furthermore, we use CLIP-Text score \cite{radford2021learning} to evaluate the instruction-following capability. 
Besides, VicoEdit is compared with baselines regarding the MS-SSIM and CLIP-Image score in Appendix \ref{sec:appendix-qualitative-results}.

\begin{figure*}[ht!]
\centering
\includegraphics[width=0.99\textwidth]{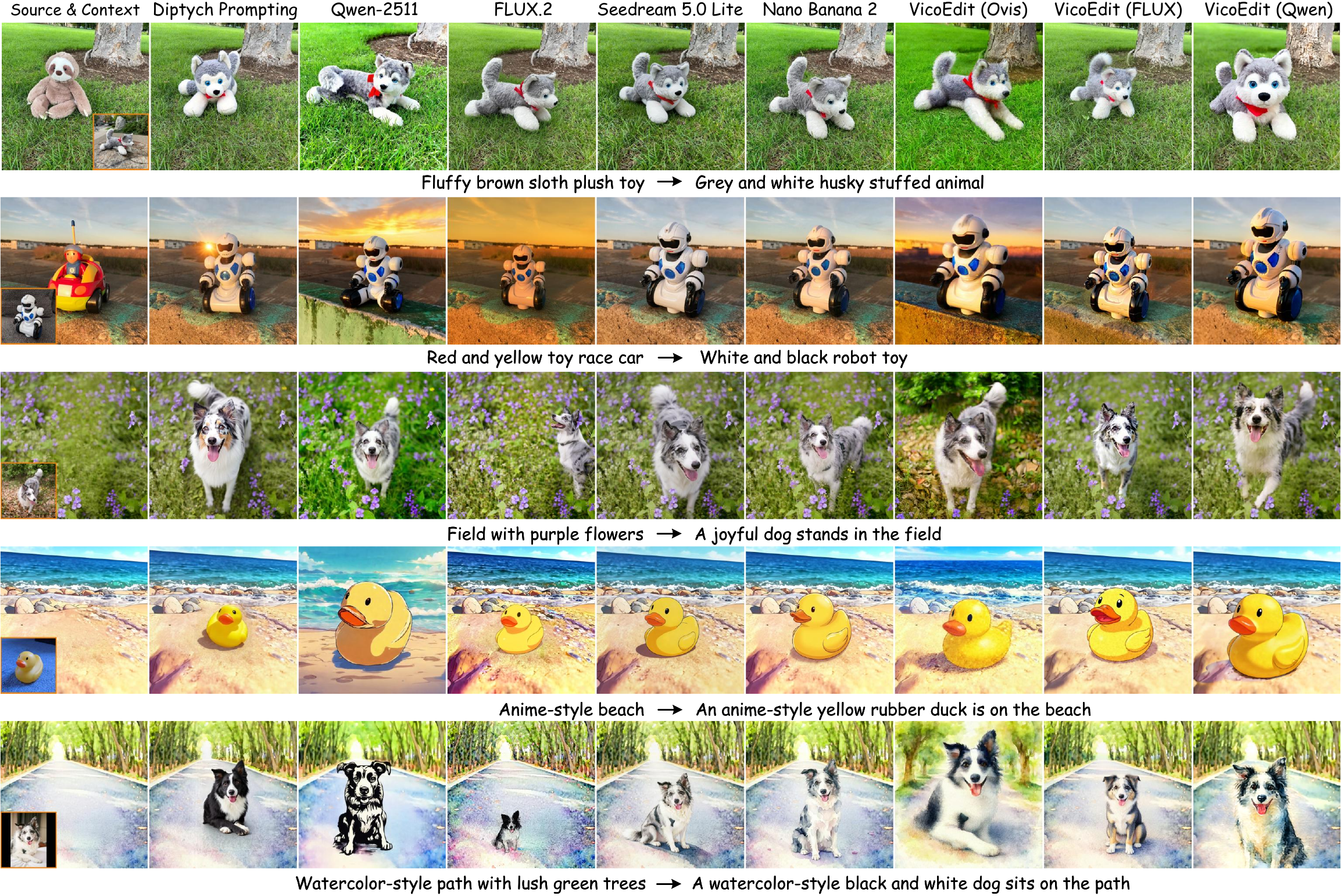}
\caption{Source image, context image, and editing results of different methods.}
\label{fig:qualitative results}
\end{figure*}

\begin{table}[t]
\centering
\setlength{\tabcolsep}{1.5pt}
\caption{Comparison on model size and editing performance. Ovis, FLUX, and Qwen are abbreviations of Ovis-U1, FLUX.1-Kontext, and Qwen-Image. The best results are marked in \textbf{bold}, while the second best results are \underline{underlined}.}
\begin{small}
\begin{tabular}{ccccc}
\bottomrule[1.2pt]
Method         & \#Params & LPIPS ($\downarrow$) & DINO ($\uparrow$) & CLIP-T ($\uparrow$) \\ \hline
Diptych Prompting  & 12B      & 0.429 & 0.734 & 0.378  \\
Qwen-2511          & 20B      & 0.521 & 0.679 & 0.381  \\ 
FLUX.2             & 32B      & \underline{0.345} & 0.683 & 0.381  \\ \hline
Seedream 5.0 Lite  & -        & 0.361 & \textbf{0.813} & \textbf{0.389} \\
Nano Banana 2      & -        & 0.378 & \underline{0.786} & \underline{0.387} \\ \hline
VicoEdit (Ovis)    & 3B       & 0.515 & 0.720 & \underline{0.387}  \\
VicoEdit (FLUX)    & 12B      & \textbf{0.322} & 0.731 & 0.378  \\
VicoEdit (Qwen)    & 20B      & 0.448 & 0.725 & 0.383  \\ \toprule[1.2pt]
\end{tabular}
\end{small}
\label{table:main-results}
\end{table}

\subsection{Main Results}
Experimental results of VicoEdit and baseline methods are shown in Table \ref{table:main-results}. Considering that Diptych Prompting is built on FLUX.1-dev \cite{flux2024}, it is natural to compare the FLUX version of VicoEdit with this training-free baseline. It is shown that VicoEdit performs better regarding structure preservation (LPIPS), and yields similar context integration (DINO similarity) and instruction following capabilities (CLIP-T score). Note that Diptych Prompting requires a user-defined mask to determine the regions to be edited. In contrast, VicoEdit is mask-free, which implicitly determines the modified areas based on the text prompt when predicting the velocity field. VicoEdit outperforms Diptych Prompting even in this more challenging setting, which demonstrates its effectiveness and flexibility.

VicoEdit achieves superior performance to the trained open-source context-aware editing models as well. The FLUX and Qwen version of VicoEdit surpass FLUX.2 and Qwen-2511 on most metrics. A reasonable explanation is that these baseline methods lack specific designs to accurately reproduce the details in the source image, whereas the proposed concept alignment strategy offers effective guidance to ensure such consistency. Furthermore, these models treat editing as a conditional generation task, and the latent $\boldsymbol{z}_t$ is initialized by random noise. In this case, the network needs to inject source image features to $\boldsymbol{z}_t$ through MMDiT blocks, where fine-grained details may be lost. In contrast, VicoEdit directly preserves these features in $\boldsymbol{z}_t$, thereby improving the editing fidelity.

Meanwhile, VicoEdit delivers comparable performance to the closed-source commercial image editing models. Specifically, the FLUX version of VicoEdit yields comparable structure preserving (LPIPS) and instruction following capabilities (CLIP-T score) to Seedream 5.0 Lite and Nano Banana 2, while the consistency to the context image (DINO similarity) is marginally lower. However, the training-free property of VicoEdit offers several unique advantages. Firstly, it circumvents the need for expensive data curation and large-scale pretraining. Furthermore, the proposed sampling algorithm offers superior interpretability compared to the conditional generation pipelines of pretrained models. Finally, VicoEdit can be leveraged to synthesize high-quality data for training multi-reference generation models at low cost.

Editing results of VicoEdit and other baseline methods are presented in Fig. \ref{fig:qualitative results}. It is shown that VicoEdit excels at maintaining features of source and context images. In particular, VicoEdit generates faithful and visually coherent results for cross-domain editing. Other methods may deviate too much from the source and context images, or fail to change the style of the subject.

Furthermore, our experiments demonstrate that VicoEdit is robust to the choice of base model. Among different base models, we observe that FLUX.1 Kontext and Qwen-Image-Edit deliver comparable performance, while Ovis-U1 slightly lags behind. This phenomenon may arise from its relatively small model size. We believe that the performance of VicoEdit can be further enhanced by leveraging a stronger base model.

\begin{table}[t]
\centering
\setlength{\tabcolsep}{1.5pt}
\caption{Sampling time and peak GPU memory usage of VicoEdit and baseline methods.}
\begin{small}
\begin{tabular}{cccccc}
\bottomrule[1.2pt]
Method        & FLUX.2 & \begin{tabular}[c]{@{}c@{}}VicoEdit \\ (FLUX)\end{tabular} & \begin{tabular}[c]{@{}c@{}}Qwen-\\ 2511\end{tabular} & \begin{tabular}[c]{@{}c@{}}VicoEdit \\ (Qwen)\end{tabular} & \begin{tabular}[c]{@{}c@{}}VicoEdit \\ (Ovis)\end{tabular} \\ \hline
Time (s)      & 193    & 122             & 83        & 184             & 81              \\
Memory (GB) & $>$80 & 55              & 59        & 70              & 30              \\ \toprule[1.2pt]
\end{tabular}
\end{small}
\label{table:complexity}
\end{table}

\subsection{Complexity Analysis}
The complexity analysis of VicoEdit and baseline methods is presented in Table \ref{table:complexity}. In particular, the memory consumption of FLUX.2 exceeds the 80GB VRAM size of the H100 GPU, and hence we have to offload its text encoder from the GPU after text embedding extraction. At the same model capacity level (e.g., Qwen-2511 versus VicoEdit Qwen), VicoEdit requires longer inference time because it needs to compute the velocity fields for $K$ times at each timestep. Meanwhile, the gradient computation for the concept alignment guidance also slightly increases the inference time and memory usage. However, the complexity of VicoEdit still remains at a practically applicable level. One may opt to deploy VicoEdit on more efficient base models (e.g, the parameter distilled or few-step distilled models), or reduce the number of sampled noise $K$ (see Appendix \ref{sec:appendix-ablation}) to further control the computation overhead.

\begin{figure*}[t!]
\centering
\includegraphics[width=0.96\textwidth]{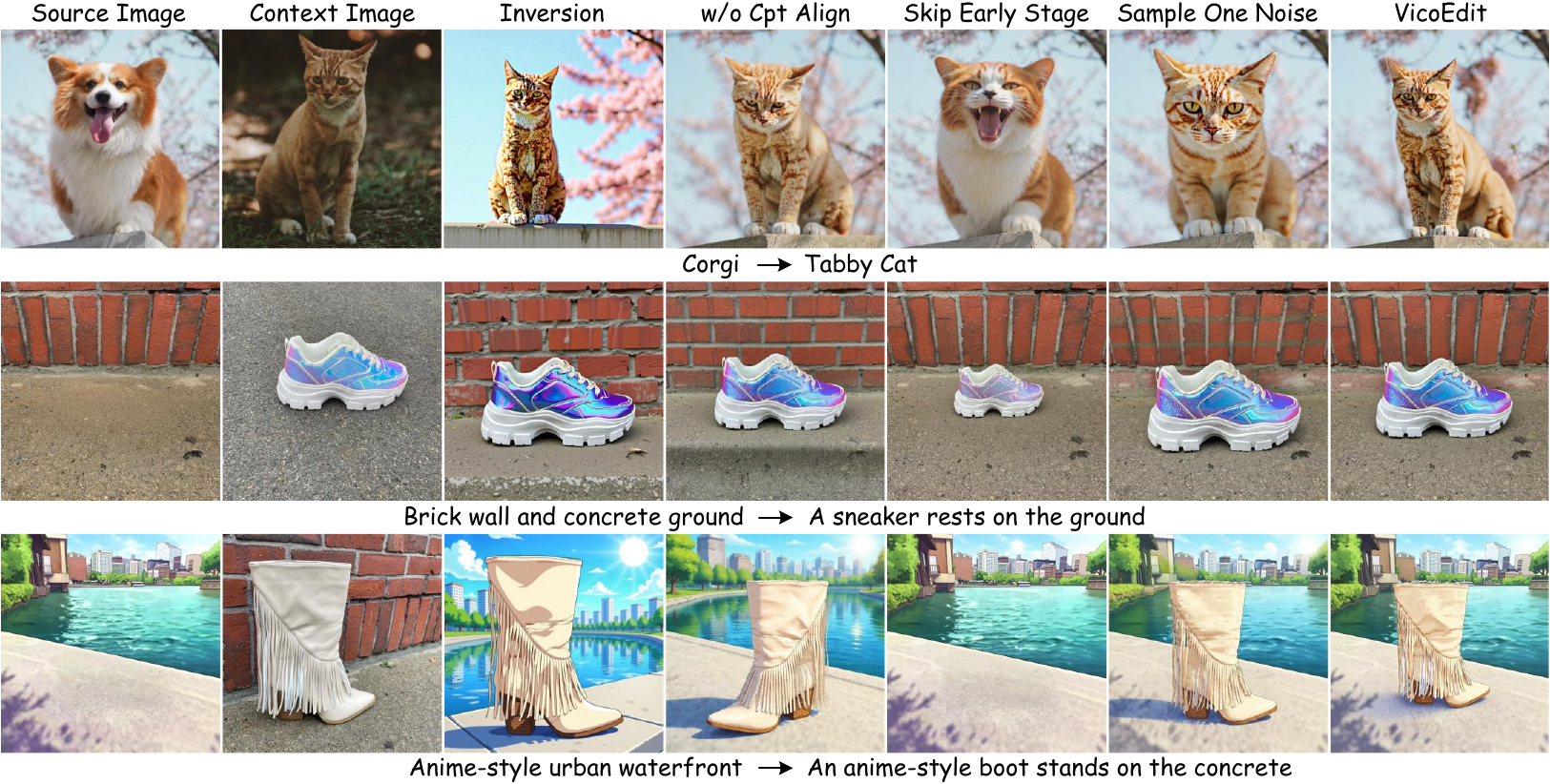}
\caption{Source, context, and edited images produced by VicoEdit and its variants.}
\label{fig:ablation results}
\end{figure*}

\subsection{Ablation Study}
In this section, we investigate the effectiveness of several key designs in VicoEdit. All experiments use FLUX.1 Kontext as the base model, and sampling takes $N=50$ steps.

\subsubsection{Inversion-Free Editing}
To investigate the strength of the inversion-free visual context integration strategy, we replace it with an inversion-based pipeline. Specifically, we use RF-Solver \cite{wang2025taming} for inversion. The sampling process initializes $\boldsymbol{z}_t$ with $\boldsymbol{z}^{src,*}$, and the visual context $\boldsymbol{z}^{ctx}$ is aggregated with $\boldsymbol{z}_t$ using the MMDiT blocks. This inversion-based model also exploits the concept alignment guidance to isolate its impact. As shown in Table \ref{table:ablation}, the inversion-free VicoEdit yields significantly better structure preserving performance (i.e., LPIPS). Fig. \ref{fig:ablation results} also shows that editing results of the inversion-based method obviously differ from the source image, while VicoEdit maintains high fidelity. These results validate the superiority of the inversion-free paradigm.

\begin{table}[t]
\centering
\setlength{\tabcolsep}{1.5pt}
\caption{Ablation study on inversion-free sampler, concept alignment, and sampling strategies.}
\begin{small}
\begin{tabular}{cccc}
\bottomrule[1.2pt]
Method              & LPIPS ($\downarrow$) & DINO ($\uparrow$)  & CLIP-T ($\uparrow$) \\ \hline
VicoEdit           & 0.322 & 0.731 & 0.378  \\
Inversion           & 0.704 & 0.707 & 0.371 \\
w/o Cpt Align        & 0.454 & 0.743 & 0.381  \\
Skip Early Stage & 0.147 & 0.427 & 0.332  \\
Sample One Noise  & 0.362 & 0.730 & 0.378  \\ \toprule[1.2pt]
\label{table:ablation}
\end{tabular}
\end{small}
\end{table}

\subsubsection{Concept Alignment}
Then, we remove the concept alignment (Cpt Align) module. The derived model updates the latent $\boldsymbol{z}_t$ using only $\tilde{\boldsymbol{v}}_t$, without introducing $\hat{\boldsymbol{v}}_t$. Table \ref{table:ablation} shows that concept alignment effectively enhances the visual consistency between the original and edited images. The improvement is further confirmed by the visualization results in Figs. \ref{fig:dps_visualization} and \ref{fig:ablation results}.

\subsubsection{Sampling Strategies}
Text-prompted editing methods commonly skip a few beginning steps to avoid significant deviations from the original image. Nonetheless, as discussed in Sec \ref{sec:visual-context-integration}, such a sampling strategy may fail to strictly follow the visual context. To demonstrate this, we test VicoEdit with $n_{\text{max}}=40$, which means that the beginning $10$ sampling steps have been skipped. In contrast, the original VicoEdit model sets $n_{\text{max}}=47$. The results in Fig. \ref{fig:ablation results} show that the model struggles to introduce the visual context when the early sampling stage is omitted, which explains the significant drops of DINO similarity and CLIP-T score in Table \ref{table:ablation}. Consequently, we opt to set the beginning timestep $t_{n_\text{max}}$ close to $1$ for context-aware editing tasks.

Additionally, we examine the impact of the number of noise samples (i.e., $K$). Table \ref{table:ablation} suggests that the consistency to the source image has been moderately enhanced by increasing $K$ from $1$ to $3$, where the latter is the default setting of VicoEdit. We also observe that the visual quality of editing results has been improved, as depicted in Fig. \ref{fig:ablation results}.

\section{Conclusion}
This paper proposes a training-free method to generalize the text-prompted editing models to context-aware editing tasks. To this end, we design an inversion-free workflow that directly translates the source image into the target one. 
Furthermore, we present a concept alignment approach to identify and preserve the unchanged regions in the source image. The proposed training-free method circumvents the high cost of data collection and large-scale pretraining required by the training-based approaches. Moreover, experiments verify that our method surpasses state-of-the-art models regarding editing consistency.

\bibliography{example_paper}
\bibliographystyle{icml2026}

\newpage
\appendix
\onecolumn

\begin{center}
	{
		\Large{\textbf{Appendix}}
	}
\end{center}

\section{Theoretical Analysis}
\subsection{Velocity Field Decomposition}
\label{sec:appendix-velocity-decomposition}
This section derives the velocity field decomposition $u_t(\boldsymbol{z}_t|\boldsymbol{y}) = u_t(\boldsymbol{z}_t) + b_t \nabla_{\boldsymbol{z}_t} \log p (\boldsymbol{y}|\boldsymbol{z}_t)$ (Eq. \ref{eq:velocity-decomposition} in the main text). As shown by \citeauthor{zheng2023guided}, the conditional velocity field $u_t(\boldsymbol{z}|\boldsymbol{y})$ in flow matching is directly related to the score function $\nabla_{\boldsymbol{z}} \log p_t(\boldsymbol{z}|\boldsymbol{y})$. Specifically, let $p_t(\boldsymbol{z}|\boldsymbol{y})$ be a Gaussian path defined by a scheduler $\alpha_t, \sigma_t$, then its generating velocity field $u_t(\boldsymbol{z}|\boldsymbol{y})$ is related to the score function $\nabla_{\boldsymbol{z}} \log p_t(\boldsymbol{z}|\boldsymbol{y})$ by:
\begin{align}
u_t(\boldsymbol{z}|\boldsymbol{y}) = a_t \boldsymbol{z} + b_t \nabla_{\boldsymbol{z}} \log p_t(\boldsymbol{z}|\boldsymbol{y}) , \quad \text{where } a_t = \frac{\dot{\alpha}_t}{\alpha_t}, \quad b_t = (\dot{\alpha}_t\sigma_t -\alpha_t\dot{\sigma}_t)\frac{\sigma_t}{\alpha_t}.
\label{eq:appendix-velocity-decomposition}
\end{align}
On the other hand, the unconditional velocity field $u_t(\boldsymbol{z})$ is related to the score function $\nabla_{\boldsymbol{z}} \log p_t(\boldsymbol{z})$ as:
\begin{align}
    u_t(\boldsymbol{z}) = a_t \boldsymbol{z} + b_t \nabla_{\boldsymbol{z}} \log p_t(\boldsymbol{z}).
\end{align}

Meanwhile, the score function can be factorized into:
\begin{align}
    u_t(\boldsymbol{z}|\boldsymbol{y}) = a_t \boldsymbol{z} + b_t \nabla_{\boldsymbol{z}} \log p_t(\boldsymbol{z}) + b_t\nabla_{\boldsymbol{z}} \log p_t(\boldsymbol{y}|\boldsymbol{z}).
\label{eq:appendix-score-of-flow}
\end{align}
Therefore, Eq. \ref{eq:appendix-score-of-flow} can be converted into:
\begin{align}
     u_t(\boldsymbol{z}|\boldsymbol{y}) = u_t(\boldsymbol{z}) + b_t \nabla_{\boldsymbol{z}} \log p_t(\boldsymbol{y}|\boldsymbol{z}).
\label{eq:appendix-classifier-guidance-for-flow}
\end{align}
In particular, the rectified flow model defines $\alpha_t = 1-t, \sigma_t = t$, and thus $b_t = -\frac{t}{1-t}$. Then, Eq. \ref{eq:appendix-classifier-guidance-for-flow} proves that the conditional velocity field can be obtained by combining the unconditional velocity field and the classifier guidance. Next, we show that the diffusion model has the identical score function as the flow matching model $u_t(\boldsymbol{z}|\boldsymbol{y})$, when they share the same noise schedule $\alpha_t, \sigma_t$. The probability flow ODE \cite{song2021score} of diffusion model is given by:
\begin{align}
    d{\boldsymbol{z}} = [f_t\boldsymbol{z}_t - \frac{1}{2}g_t^2 \nabla_{\boldsymbol{z}_t} \log p(\boldsymbol{z}_t|\boldsymbol{y})]dt,
\label{eq:appendix-conditional-diff-ode}
\end{align}
where $f_t = \frac{d \log \alpha_t}{dt}$, $g_t^2 = \frac{d \sigma^2_t}{dt} - 2 \frac{d \log \alpha_t}{dt} \sigma_t^2$ \cite{kingma2021variational, zheng2023guided}. These coefficients can be transferred into:
\begin{align}
    f_t = \frac{d \log \alpha_t}{dt} = \frac{\dot{\alpha}_t}{\alpha_t}, \quad
    -\frac{1}{2}g^2_t = -\frac{1}{2}\frac{d \sigma^2_t}{dt} + \frac{d \log \alpha_t}{dt} \sigma_t^2 = - \dot{\sigma}_t\sigma_t + \frac{\dot{\alpha_t}}{\alpha_t}\sigma_t^2 = (\dot{\alpha}_t\sigma_t - \alpha_t \dot{\sigma}_t)\frac{\sigma_t}{\alpha_t}.
\label{eq:appendix-coeef-of-diffusion}
\end{align}
Comparing Eq. \ref{eq:appendix-velocity-decomposition} and Eq. \ref{eq:appendix-coeef-of-diffusion}, we get $f_t = a_t, -\frac{1}{2}g_t^2 = b_t$. Therefore, the ODE solution derived from conditional flow matching (Eq. \ref{eq:appendix-score-of-flow}) coincides with the one given by the conditional diffusion model (Eq. \ref{eq:appendix-conditional-diff-ode}). According to \citeauthor{song2021score}, Eq. \ref{eq:appendix-conditional-diff-ode} leads to samples from $p(\boldsymbol{z}_0|\boldsymbol{y})$ when $t=0$. Thus, conditional flow matching (Eq. \ref{eq:appendix-score-of-flow}) also generate samples from $p(\boldsymbol{z}_0|\boldsymbol{y})$ at $t=0$. Finally, Eq. \ref{eq:appendix-classifier-guidance-for-flow} proves that combining the unconditional velocity field and the classifier guidance reaches $p(\boldsymbol{z}_0|\boldsymbol{y})$ at $t=0$ as well.

\subsection{Diffusion Posterior Sampling}
\label{sec:appendix-dps}
This section introduces the diffusion posterior sampling (DPS) \cite{chung2023diffusion}. DPS aims to estimate the image $\boldsymbol{x}_0$ based on its partial measurement $\boldsymbol{y}$, which is generated by a forward model $\mathcal{A}(\cdot)$:
\begin{align}
    \boldsymbol{y} = \mathcal{A}(\boldsymbol{x}_0) + \boldsymbol{\epsilon}, \quad \text{where } \boldsymbol{\epsilon} \in \mathcal{N}(\boldsymbol{0}, \sigma^2 \boldsymbol{I}).
\label{eq:appendix-dps-forward}
\end{align}
For example, DPS serves as an image super-resolution model when $\mathcal{A}$ is the down-sampling function. Then, $\boldsymbol{x}_0$ can be estimated using a conditional diffusion model, and the corresponding score function is:
\begin{align}
    \nabla_{\boldsymbol{x}_t} \log p(\boldsymbol{x}_t|\boldsymbol{y}) = \nabla_{\boldsymbol{x}_t} \log p(\boldsymbol{x}_t) + \nabla_{\boldsymbol{x}_t} \log p(\boldsymbol{y}|\boldsymbol{x}_t).
\label{eq:appendix-dps-score}
\end{align}
Suppose we have a trained unconditional diffusion model that can predict $\nabla_{\boldsymbol{x}_t} \log p(\boldsymbol{x}_t)$, and hence our subsequent target is to estimate $\nabla_{\boldsymbol{x}_t} \log p(\boldsymbol{y}|\boldsymbol{x}_t)$. The likelihood function $p(\boldsymbol{y}|\boldsymbol{x}_t)$ is factorized into:
\begin{align}
    p(\boldsymbol{y}|\boldsymbol{x}_t) &= \int p(\boldsymbol{y}|\boldsymbol{x}_0, \boldsymbol{x}_t) p(\boldsymbol{x}_0|\boldsymbol{x}_t) d\boldsymbol{x}_0 \notag \\
    &= \int p(\boldsymbol{y}|\boldsymbol{x}_0) p(\boldsymbol{x}_0|\boldsymbol{x}_t) d \boldsymbol{x}_0 \notag \\
    &= \mathbb{E}_{\boldsymbol{x}_0 \sim p(\boldsymbol{x}_0| \boldsymbol{x}_t)}[p(\boldsymbol{y}|\boldsymbol{x}_0)].
\label{eq:appendix-dps-likelihood}
\end{align}
The second equality is due to the conditional independence of $\boldsymbol{y}$ and $\boldsymbol{x}_t$ given $\boldsymbol{x}_0$. Then, DPS approximates the above expectation as:
\begin{align}
    \mathbb{E}_{\boldsymbol{x}_0 \sim p(\boldsymbol{x}_0| \boldsymbol{x}_t)}[p(\boldsymbol{y}|\boldsymbol{x}_0)] \approx p(\boldsymbol{y}|\hat{\boldsymbol{x}}_0), \quad \text{where } \hat{\boldsymbol{x}}_0 := \mathbb{E}[\boldsymbol{x}_0 | \boldsymbol{x}_t].
\label{eq:appendix-dps-approximation}
\end{align}
For the diffusion models that are built on the variance-preserving forward process, $p(\boldsymbol{x}_t | \boldsymbol{x}_0)$ is a Gaussian distribution: 
\begin{align}
    p(\boldsymbol{x}_t | \boldsymbol{x}_0) \sim \mathcal{N}(\sqrt{\bar{\alpha}_t} \boldsymbol{x}_0, (1 - \bar{\alpha}_t) \boldsymbol{I}).
\end{align}

Thus, the expectation $\mathbb{E}[\boldsymbol{x}_0 | \boldsymbol{x}_t]$ can be obtained using Tweedie's formula:
\begin{align}
    \hat{\boldsymbol{x}}_0 = \mathbb{E}[\boldsymbol{x}_0 | \boldsymbol{x}_t] = \frac{1}{\sqrt{\bar{\alpha}_t}} (\boldsymbol{x}_t + (1 - \bar{\alpha}_t) \nabla_{\boldsymbol{x}_t} \log p(\boldsymbol{x}_t)).
\label{eq:appendix-dps-tweedie}
\end{align}

The definition in Eq. \ref{eq:appendix-dps-forward} implies that $p(\boldsymbol{y}|\boldsymbol{x}_0) \sim \mathcal{N}(\mathcal{A}(\boldsymbol{x}_0), \sigma^2\boldsymbol{I})$, and the corresponding probability density function is:
\begin{align}
   p(\boldsymbol{y}|\boldsymbol{x}_0) = \frac{1}{\sqrt{(2\pi)^n \sigma^{2n}}} \text{exp}[-\frac{||\boldsymbol{y} - \mathcal{A}(\boldsymbol{x}_0)||^2_2}{2\sigma^2}],
\label{eq:appendix-dps-pdf}
\end{align}
where $n$ is the dimension of $\boldsymbol{y}$. Combining Eq. \ref{eq:appendix-dps-pdf} and Eq. \ref{eq:appendix-dps-approximation}, $\nabla_{\boldsymbol{x}_t} \log p(\boldsymbol{y}|\boldsymbol{x}_t)$ is given by:
\begin{align}
    \nabla_{\boldsymbol{x}_t} \log p(\boldsymbol{y}|\boldsymbol{x}_t) \approx
    \nabla_{\boldsymbol{x}_t} \log p(\boldsymbol{y}|\hat{\boldsymbol{x}}_0) = -\frac{1}{\sigma^2} \nabla_{\boldsymbol{x}_t}||\boldsymbol{y} - \mathcal{A}(\hat{\boldsymbol{x}}_0)||^2_2.
\label{eq:appendix-dps-result}
\end{align}
By taking Eq. \ref{eq:appendix-dps-tweedie} into Eq. \ref{eq:appendix-dps-result}, we get the approximation of $\nabla_{\boldsymbol{x}_t} \log p(\boldsymbol{y}|\boldsymbol{x}_t)$. Finally, the score function $\nabla_{\boldsymbol{x}_t} \log p(\boldsymbol{x}_t|\boldsymbol{y})$ is estimated by $\nabla_{\boldsymbol{x}_t} \log p(\boldsymbol{x}_t) + \nabla_{\boldsymbol{x}_t} \log p(\boldsymbol{y}|\hat{\boldsymbol{x}}_0)$.

\section{Implementation Details}
\subsection{VicoEdit Settings}
\label{sec:appendix-vicoedit-settings}
Table \ref{table:hyper-parameters} specifies the hyper-parameter settings of VicoEdit under different base models. Following FlowEdit, we use different classifier-free guidance (CFG) scales to individually modulate the inverse and sampling velocities.
For Qwen-Image-Edit and Ovis-U1, we apply CFG on both text and image modalities as:
\begin{align}
    \boldsymbol{v}(\boldsymbol{z}^{tar}_t, \boldsymbol{r}, \boldsymbol{z}^{ctx}) = \boldsymbol{f}(\boldsymbol{z}^{tar}_t, \varnothing, \varnothing) + c_{I} (\boldsymbol{f}(\boldsymbol{z}^{tar}_t, \varnothing, \boldsymbol{z}^{ctx}) - \boldsymbol{f}(\boldsymbol{z}^{tar}_t, \varnothing, \varnothing)) + c_{T} (\boldsymbol{f}(\boldsymbol{z}^{tar}_t, \boldsymbol{r}, \boldsymbol{z}^{ctx}) - \boldsymbol{f}(\boldsymbol{z}^{tar}_t, \varnothing, \boldsymbol{z}^{ctx})).
\end{align}
Since FLUX has performed CFG distillation, we apply a unified CFG scale for both modalities. Our evaluation dataset comprises three tasks: in-domain replacement, in-domain add, and cross-domain add. To achieve the best performance, we opt to enhance the CFG guidance for in-domain add and cross-domain add tasks. Other hyper-parameters are kept the same for all tasks. Besides, following the common practice in DPS, we set the controlling scale $\alpha_t$ of the concept-guided posterior sampling as a time-independent constant. 

\begin{table}[h]
\centering
\caption{Hyper-parameters of VicoEdit. $c^{tar}_T$ and $c^{tar}_I$ columns showcase the CFG scales for replacement and add tasks.}
\begin{small}
\begin{tabular}{cccccccccc}
\bottomrule[1.2pt]
Model & $N$ & $n_\text{max}$ & $K$ & $c^{src}_T$ & $c^{tar}_T$ & $c^{tar}_I$ & $\tau$ & $\alpha_t$ & $\sigma$ \\ \hline
FLUX.1-Kontext  & 50  & 47             & 3   & 1.5              & 5.5 / 6.0        & -         & 0.25   & 0.5      & 0        \\
Qwen-Image-Edit  & 50  & 45             & 2   & 1.5              & 7.5 / 8.0        & 2.5 / 3.0 & 0.25   & 1.0      & 0        \\
Ovis-U1  & 50  & 47             & 3   & 1.5              & 6.5 / 7.0        & 2.5 / 3.0 & 0.25   & 1.0      & 0        \\ \toprule[1.2pt]
\end{tabular}
\end{small}
\label{table:hyper-parameters}
\end{table}

\subsection{Prompt Template for Baseline Models}
\label{sec:appendix-baseline-settings}
This section details the prompts utilized to evaluate FLUX.2, Qwen-Image-Edit-2511, Seedream 5.0 Lite, and Nano Banana 2. We use a template that combines editing instructions with image captions, as our experiments indicate this approach outperforms using either the instruction or the caption alone. The templates for different tasks are listed below:

\begin{itemize}
\item \textbf{In-domain replacement}: Replace the \texttt{\{source subject\}} in the first image with the \texttt{\{context subject\}} in the second image to generate such an image: ``\texttt{\{target image caption\}}''.

\item \textbf{In-domain add}: Add the \texttt{\{context subject\}} in the second image to the first image to generate such an image: ``\texttt{\{target image caption\}}''.

\item \textbf{Cross-domain add}: Transfer the \texttt{\{context subject\}} in the second image into the \texttt{\{style\}} style. Then insert it to the first image to generate such an image: ``\texttt{\{target image caption\}}''.

\end{itemize}

\subsection{Evaluation Dataset}
\label{sec:appendix-dataset}
Our dataset is built upon the DreamBooth dataset, which provides images containing a single primary subject alongside its class label, \texttt{subject}. Each test sample consists of a source image $\boldsymbol{x}^{src}$, its caption $\boldsymbol{c}^{src}$, a context image $\boldsymbol{x}^{ctx}$, a target caption of the edited image $\boldsymbol{c}^{tar}$, preserved concepts $\boldsymbol{c}^{cpt}_{pos}$, and changed concepts $\boldsymbol{c}^{cpt}_{neg}$.

We employ the following pipeline for data curation. First, we manually match the images in the DreamBooth dataset to form reasonable source-context image pairs. Next, we adopt FLUX.2 to process the source image: for the in-domain add task, \texttt{subject} in the image is removed to generate the background; for the cross-domain add task, we extract the background and subsequently transfer its style; for the in-domain replacement task, the source image remains unchanged. Following this, we employ Gemini 2.5 Flash to generate the caption $\boldsymbol{c}^{src}$ of the processed source image $\boldsymbol{x}^{src}$. We then feed both $\boldsymbol{x}^{src}$ and $\boldsymbol{x}^{ctx}$ to Gemini 2.5 Flash to produce the caption $\boldsymbol{c}^{tar}$ tailored to the specific task requirements. Finally, we utilize Gemini 2.5 Flash to identify the two most salient concepts in the source image's background, which correspond to the preserved concepts $\boldsymbol{c}^{cpt}_{pos}$. The changed concepts $\boldsymbol{c}^{cpt}_{neg}$ include \texttt{source subject} and \texttt{target subject} for the in-domain replacement task, whereas comprising only \texttt{source subject} for in-domain and cross-domain add tasks.

\section{Additional Results}
\subsection{Evaluation on More Metrics}
\label{sec:appendix-quantitative-results}
In the main text, we use LPIPS and DINO similarity to measure the editing consistency. Here, we further evaluate the editing fidelity using other metrics. Specifically, we compare the structure similarity index (SSIM) between the source image and the editing result. Besides, in addition to the DINO feature space, this section further compares the similarity between the subjects in context and edited images based on the CLIP embedding. As shown in Table \ref{table:appendix-ssim}, VicoEdit outperforms other open-source training-free or training-based editing methods regarding SSIM and CLIP similarity as well. These results demonstrate the strength of VicoEdit in preserving the structure in source and context images.

\begin{table*}[t]
\setlength{\tabcolsep}{1.5pt}
\centering
\caption{Comparisons on SSIM and CLIP similarity. Diptych Pmt. denotes Diptych Prompting. FLUX, Qwen, Ovis represent the FLUX.1-Kontext, Qwen-Image-Edit, and Ovis-U1 version of VicoEdit. The best results are marked in \textbf{bold}, while the second best ones are \underline{underlined}.}
\begin{small}
\begin{tabular}{ccccccccc}
\bottomrule[1.2pt]
Metric & Diptych Pmt. & FLUX.2 & Qwen-2511 & Seedream 5.0 Lite & Nano Banana 2 & VicoEdit (FLUX) & VicoEdit (Qwen) & VicoEdit (Ovis) \\ \hline
SSIM $(\uparrow)$   & 0.346        & \underline{0.657}  & 0.451     & 0.652    & 0.605       & \textbf{0.744}           & 0.611           & 0.488           \\
CLIP-I $(\uparrow)$ & 0.899        & 0.873  & 0.881     & \textbf{0.941}    & \underline{0.931}       & 0.909           & 0.896           & 0.905           \\ 
\toprule[1.2pt]
\end{tabular}
\end{small}
\label{table:appendix-ssim}
\end{table*}

\begin{table*}[ht]
\centering
\caption{Performance of VicoEdit under different hyper-parameter settings.}
\begin{tabular}{cccccccc}
\bottomrule[1.2pt]
Metric & VicoEdit & $K=2$ & $K=1$ & $\tau=0.5$ & $\tau=0.1$ & $\alpha_t=0.25$ & $\alpha_t=1$ \\ \hline
LPIPS ($\downarrow$)  & 0.322    & 0.338 & 0.362 & 0.289      & 0.401      & 0.425           & 0.358        \\
DINO ($\uparrow$)   & 0.731    & 0.729 & 0.730 & 0.666      & 0.730      & 0.735           & 0.733        \\
CLIP-T ($\uparrow$) & 0.378    & 0.379 & 0.378 & 0.370      & 0.379      & 0.380           & 0.380        \\ \toprule[1.2pt]
\end{tabular}
\label{table:appendix-ablation}
\end{table*}

\subsection{Impact of Hyper-Parameters}
\label{sec:appendix-ablation}
This section analyzes the impact of several key hyper-parameters, specifically $K$, $\alpha_t$, and $\tau$. In VicoEdit, $K$ determines the number of noise samples used to estimate the velocity field $\tilde{\boldsymbol{v}}$. As shown in Table \ref{table:appendix-ablation}, increasing $K$ leads to more precise estimation of $\tilde{\boldsymbol{v}}$, thereby improving overall performance. Meanwhile, the performance of VicoEdit is quite close when $K$ is set to $2$ or $3$. In the main experiment, we utilize $K = 3$ for the best performance, but it is also feasible to reduce $K$ to 2 for faster sampling. This configuration shortens the inference time from $\mathbf{122}$s to $\mathbf{89}$s with minor performance degradation.

The parameter $\tau$ serves as the threshold for designating tokens as preserved concepts. A lower $\tau$ results in a larger number of tokens being classified as preserved, thereby constraining the scope of concept alignment guidance. Conversely, a higher $\tau$ extends this guidance to broader regions but simultaneously heightens the risk of misclassification. As shown in Table \ref{table:appendix-ablation}, an excessively low $\tau$ (e.g., $\tau = 0.1$) renders the guidance too restrictive, leading to a performance decline in LPIPS. On the other hand, an overly large $\tau$ (e.g., $\tau = 0.5$) may apply guidance to modified regions, which compromises subject consistency and instruction following capabilities (i.e., DINO and CLIP-T scores). Empirically, we found an appropriate setting of $\tau$ is $0.25$. 

Finally, the parameter $\alpha_t$ modulates the strength of the concept alignment guidance. Results in Table \ref{table:appendix-ablation} reveal that insufficient guidance (e.g., $\alpha_t=0.25$) fails to adequately preserve the concepts in the unmodified regions, which explains the performance decline of LPIPS. Meanwhile, an inappropriately large guidance also influences the performance, because it disturbs the guidance of the semantic velocity field $\tilde{\boldsymbol{v}}_t$.

\subsection{Qualitative Results}
\label{sec:appendix-qualitative-results}
\begin{figure*}[ht]
\centering
\includegraphics[width=0.99\textwidth]{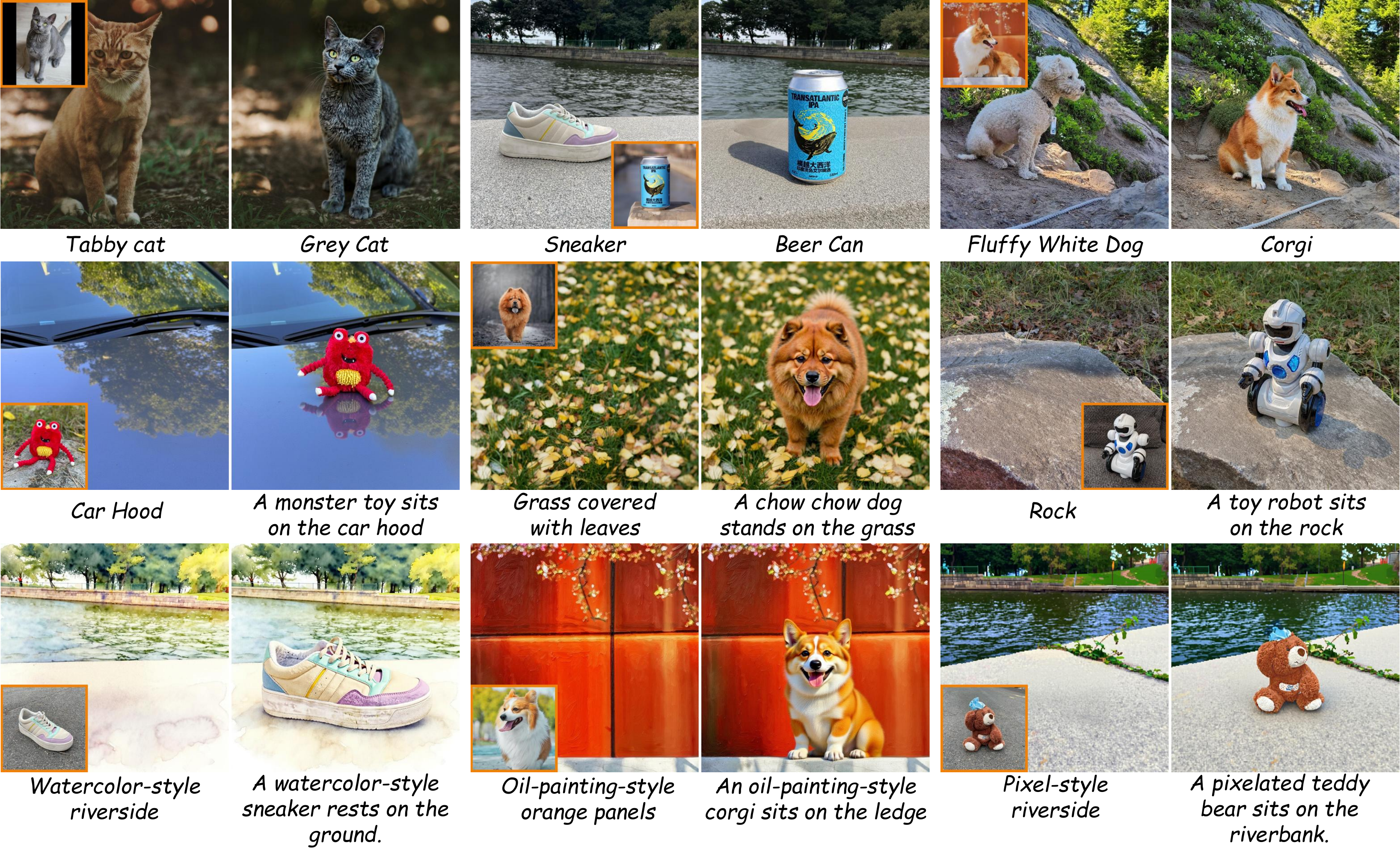}
\caption{Results produced by FLUX.1-Kontext. The left half of each image pair shows the source and context images, while the right half presents the edited image.}
\label{fig:appendix-flux}
\end{figure*}

\begin{figure*}[ht]
\centering
\includegraphics[width=0.99\textwidth]{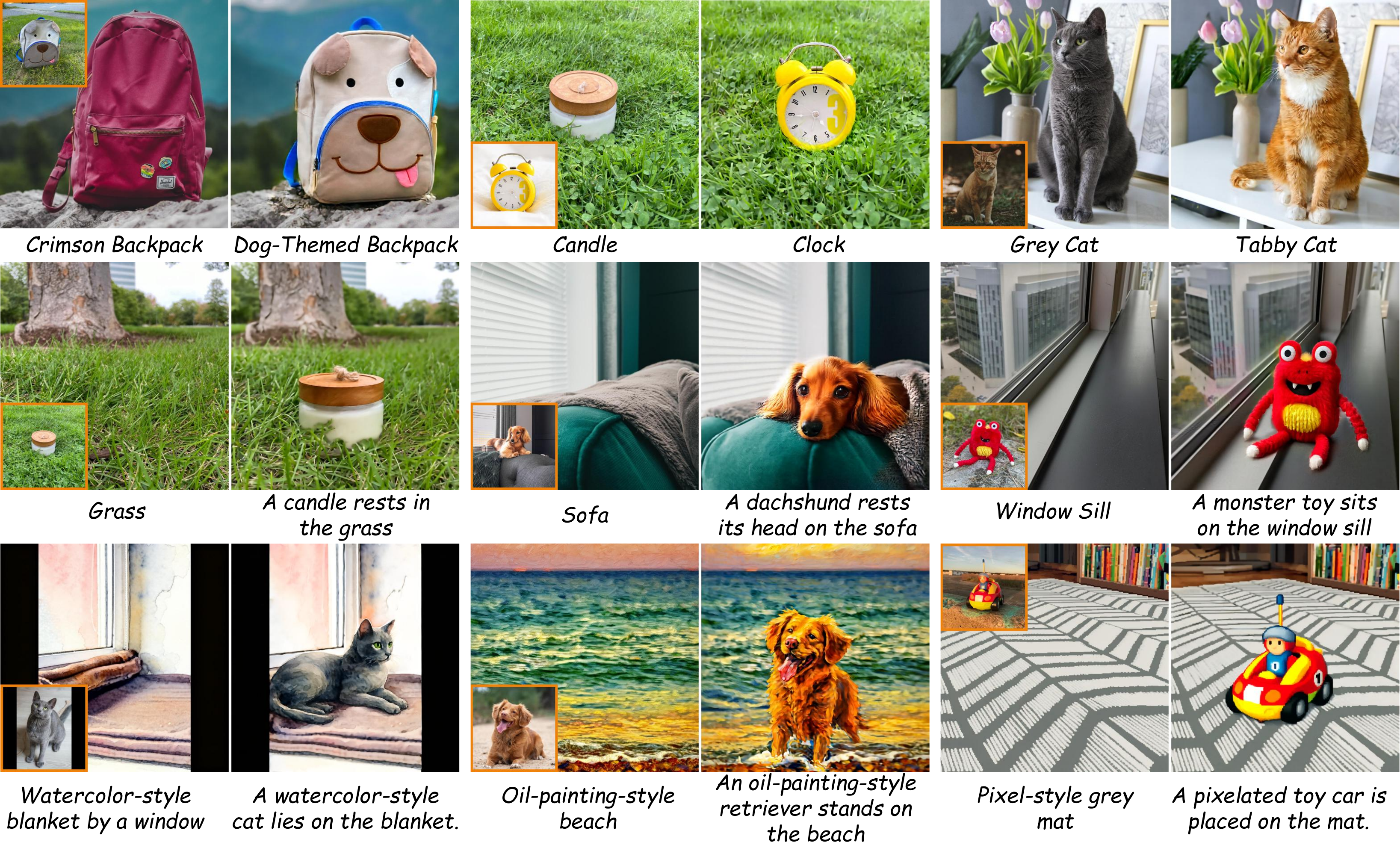}
\caption{Results generated by Qwen-Image-Edit. The left half of each pair shows the source and context images, while the right half corresponds to the editing result.}
\label{fig:appendix-qwen}
\end{figure*}

\begin{figure*}[ht]
\centering
\includegraphics[width=0.99\textwidth]{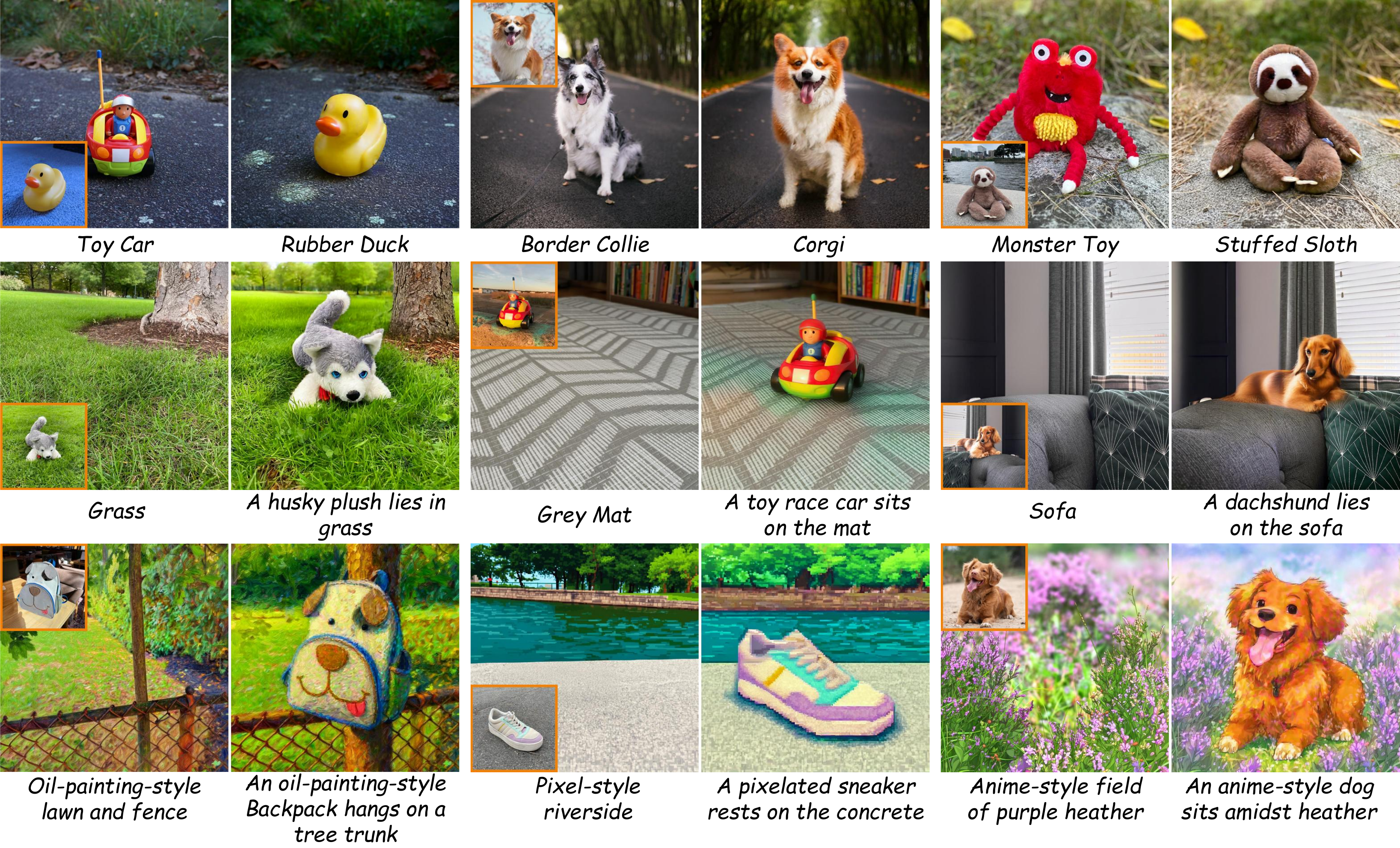}
\caption{Editing results of Ovis-U1. The left column of each pair exhibits the source and context images, while the right column shows the editing result.}
\label{fig:appendix-ovis}
\end{figure*}

This section presents more editing results produced by VicoEdit. The images generated by FLUX.1-Kontext, Qwen-Image-Edit, and Ovis-U1 are exhibited in Fig. \ref{fig:appendix-flux}, Fig. \ref{fig:appendix-qwen}, and Fig. \ref{fig:appendix-ovis}, respectively. These results demonstrate that VicoEdit can generate coherent and visually appealing pictures, while preserving the detailed patterns in source and context images.

\end{document}